%% file: main.tex
\documentclass[sigconf]{acmart}
\usepackage{subcaption}
\usepackage{multirow}
\usepackage{enumitem}
\usepackage{tabularx}
\newcolumntype{C}{>{\centering\arraybackslash}X}
\AtBeginDocument{%
  }

\setcopyright{acmlicensed}
\copyrightyear{2026}
\acmYear{2026}
\acmDOI{XXXXXXX.XXXXXXX}
\acmConference[Conference acronym 'XX]{Make sure to enter the correct
  conference title from your rights confirmation email}{June 03--05,
  2018}{Woodstock, NY}
\acmISBN{978-1-4503-XXXX-X/2018/06}

\begin{document}

\title{Learning to Transfer Across Modes: Towards Unified Urban Mobility Forecasting}


\author{Yixuan Zhao}
\affiliation{%
  \institution{University of Exeter}
  \department{Department of Computer Science}
  \city{Exeter}
  \postcode{EX4 4QF}
  \country{United Kingdom}
}
\email{yz776@exeter.ac.uk}

\author{Man Luo}
\authornote{Corresponding author.}
\affiliation{%
  \institution{University of Exeter}
  \department{Department of Computer Science}
  \city{Exeter}
  \postcode{EX4 4QF}
  \country{United Kingdom}
}
\email{m.luo@exeter.ac.uk}

\renewcommand{\shortauthors}{Trovato et al.}

\begin{abstract}

Urban transportation systems consist of multiple mobility modes that coexist within the same city and exhibit complex interdependencies, leading to correlated demand dynamics across modes. However, forecasting demand jointly across different modes remains challenging due to substantial heterogeneity in space and the limited availability of historical data for emerging modes. Existing forecasting methods are largely developed for individual mobility modes and implicitly assume compatible spatial structures between source and target systems, which severely restricts their applicability in multi-modal settings. To address these challenges, we propose \textbf{TransMod}, a unified framework for urban mobility demand forecasting that enables effective knowledge transfer across heterogeneous mobility modes. TransMod constructs a shared zone-level spatial representation that aligns mobility systems with different spatial granularities into a common space, thereby reducing structural mismatch and distributional shift. Built on this unified representation, TransMod further learns transferable spatio-temporal patterns from data-rich source modes and adapts them to data-scarce target modes, alleviating the dependence on extensive target-domain histories. Extensive experiments on real-world datasets demonstrate that TransMod consistently outperforms existing approaches and provides robust forecasting performance under limited target data.

\end{abstract}



\maketitle

\input{00_Introduction}
\input{01_Preliminaries}
\input{02_Methodology}
\input{03_Experiments}
\input{04_Related_Work}
\input{05_Conclusion}

\bibliographystyle{ACM-Reference-Format}
\bibliography{reference}

\end{document}

%% file: 00_Introduction.tex
\section{Introduction}

Urban mobility increasingly operates as an ecosystem of multiple transportation modes that coexist and interact within the same city, such as bike-sharing, metro, and ride-hailing systems~\cite{10.7551/mitpress/9399.001.0001}. Despite differences in service mechanisms, infrastructure, and operational characteristics, these modes are driven by shared underlying human mobility demand, giving rise to strongly correlated and temporally synchronized demand dynamics across modes~\cite{song2010limits,tang2015uncovering}. For example, commuting peaks, weather changes, and major urban events may simultaneously affect multiple mobility services, although their responses can vary across spatial scales, service types, and usage contexts. Accurately forecasting such multi-modal demand is therefore critical for urban planning, resource allocation, and operational management~\cite{vlahogianni2014short}. However, most existing forecasting models are developed for individual mobility modes and treat different systems in isolation, which overlooks the transferable demand patterns embedded across modes. This limitation reduces their ability to jointly model heterogeneous mobility modes and exploit cross-modal dependencies in realistic large-scale urban environments~\cite{ke2017short}.

\begin{figure}[tb]
    \centering
    \begin{minipage}[b]{.15\textwidth}
        \centering
        \includegraphics[width=\linewidth]{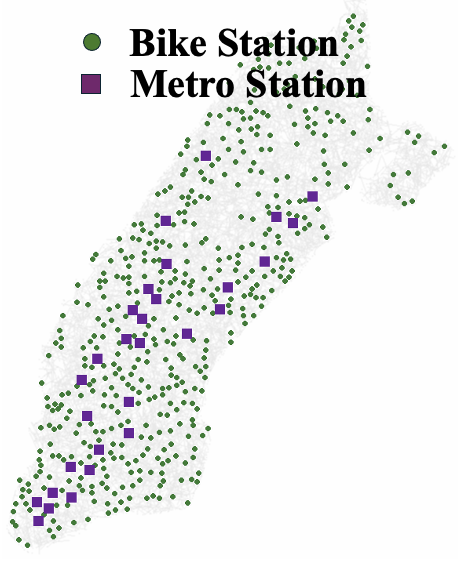}
        \subcaption{}
        \label{fig:subfig_a1}
    \end{minipage}
    \hfill
    \begin{minipage}[b]{.15\textwidth}
        \centering
        \includegraphics[width=\linewidth]{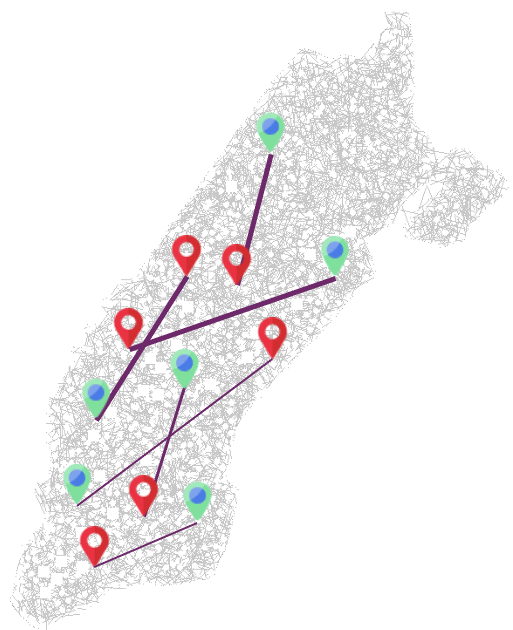}
        \subcaption{}
        \label{fig:subfig_b1}
    \end{minipage}
    \hfill
    \begin{minipage}[b]{.15\textwidth}
        \centering
        \includegraphics[width=\linewidth]{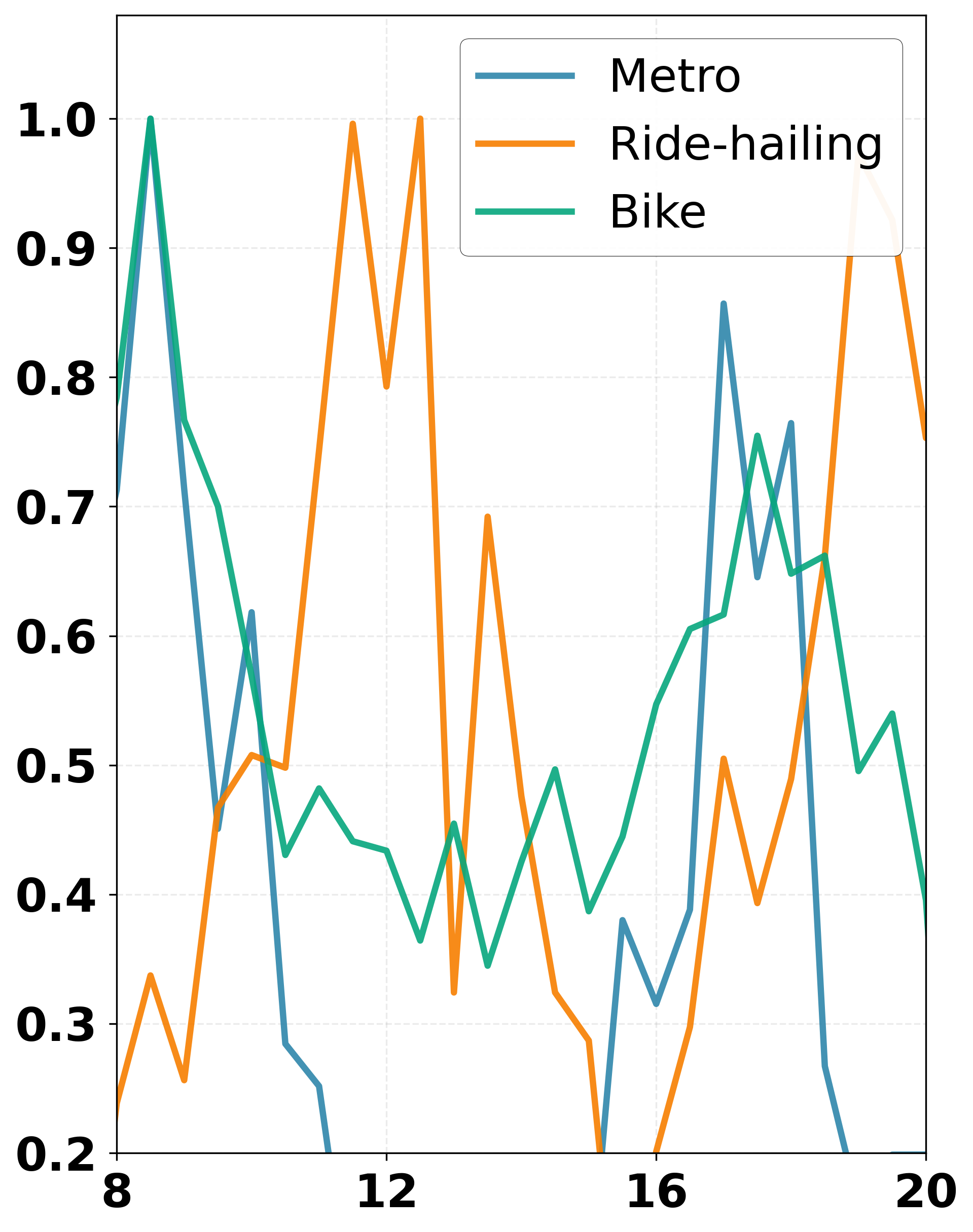}
        \subcaption{}
        \label{fig:subfig_c1}
    \end{minipage}
    \caption{Spatial and temporal heterogeneity across urban mobility modes in NYC. (a) Station-level distribution of bike-sharing and metro systems. (b) Examples of OD flow patterns of ride-hailing demand. (c) Temporal demand patterns of different systems.}
    \label{fig:my_label1}
    \Description{}
\end{figure}

Coexisting mobility modes within the same city exhibit pronounced structural heterogeneity in both spatial representation and temporal observability. Station-based shared mobility systems typically provide fine-grained spatial observations together with rich historical temporal signals~\cite{10.5555/3298239.3298479,10.5555/3504035.3504351}. In contrast, station-less or emerging mobility services record demand as individual trip events that are continuously distributed across urban space rather than anchored to fixed locations, requiring aggregation into coarser spatial units for modeling and often resulting in limited and noisy historical observations. As illustrated in Figure~\ref{fig:my_label1}(\subref{fig:subfig_a1}) and (\subref{fig:subfig_b1}), station-based systems such as bike-sharing and metro rely on discrete station-level representations that support demand forecasting at fixed spatial anchors, whereas ride-hailing operates as a station-less system characterized by spatially dispersed origin–destination flows. Figure~\ref{fig:my_label1}(\subref{fig:subfig_c1}) further highlights pronounced temporal heterogeneity across mobility modes, with metro, ride-hailing, and bike-sharing exhibiting distinct demand dynamics and highly uneven observation densities. The coexistence of mismatched spatial granularity and uneven temporal observability violates the assumptions underlying most existing forecasting and transfer learning methods, thereby limiting effective cross-system knowledge transfer~\cite{yao2019learning,jin2023transferable}. Consequently, conventional time-series and spatio-temporal graph models struggle with temporally sparse target systems due to their reliance on dense historical observations~\cite{10.5555/3367243.3367301,Geng_Li_Wang_Zhang_Yang_Ye_Liu_2019}.

Most existing urban demand forecasting approaches are designed for individual mobility systems or assume comparable data structures and observation patterns across domains. Recent transfer learning studies have attempted to bridge regions or systems by exploiting spatial correspondences and shared mobility patterns~\cite{10.1145/3447548.3467330,Zhang_Wang_Yu_Sun_Wang_Wang_2025}. While effective in conventional cross-region or cross-system settings, these methods remain insufficient for shared urban mobility environments where multiple heterogeneous modes coexist, interact, and generate demand at different spatial and temporal scales. In such multi-modal settings, effective transfer is hindered by two interrelated challenges: \textit{i)} heterogeneous mobility modes operate at incompatible spatial granularities, such that fine-grained station-level signals cannot be directly mapped to coarser region-based representations without distorting spatial interactions; and \textit{ii)} most transfer-based models require sufficient historical observations from the target system to adapt transferred dynamics, which is rarely available for emerging or data-scarce mobility services. These limitations make it difficult to establish a unified representation space for transferring knowledge from data-rich source modes to data-scarce target modes. Consequently, existing methods struggle to fully exploit heterogeneous mobility data for demand forecasting across mobility modes.

To address these challenges, we propose \textbf{TransMod}, a unified urban mobility forecasting framework that enables knowledge transfer across mobility modes. The framework is motivated by the observation that shared mobility systems coexist within the same urban space while exhibiting fundamental differences in spatial representation and temporal data availability. TransMod resolves structural heterogeneity by constructing a spatially grounded soft assignment that aggregates fine-grained station-level information into zone-level representations, establishing a unified spatial foundation for modeling across mobility modes. This assignment mechanism preserves local spatial proximity and integrates heterogeneous mobility signals in a flexible manner. Within this unified zone-level representation, the framework aligns zones across mobility modes to mitigate distributional discrepancies and further incorporates a memory-based component that leverages learned spatio-temporal dynamics from source systems with longer observation histories to inform forecasting for target systems with limited historical data. This design enables temporal knowledge transfer without requiring historical observations from the target system. Extensive experiments on real-world shared mobility systems demonstrate the effectiveness of the proposed framework. The contributions of our work are summarized as follows:

\begin{itemize}

\item To reconcile differences in spatial granularities across mobility modes, we formulate heterogeneous mobility demand forecasting as a unified zone-level prediction problem within a shared spatial framework.

\item We introduce a spatial aggregation scheme that consolidates fine-grained station representations into a unified zone-based structure while preserving spatial continuity in dynamic urban environments.

\item We introduce a transfer paradigm that enables demand forecasting across mobility modes without relying on historical temporal observations from the target system.

\item Extensive experiments on real-world shared mobility datasets demonstrate that TransMod consistently outperforms strong baselines and achieves state-of-the-art performance in zone-level demand forecasting.

\end{itemize}

%% file: 01_Preliminaries.tex
\section{Preliminaries}

We consider an urban area served by multiple heterogeneous shared mobility systems, including station-based systems such as bike-sharing and metro, where demand is recorded at fixed geographic anchors, and a zone-based system such as ride-hailing, where trip demand is aggregated over spatial zones. These systems differ fundamentally in spatial granularity and temporal data availability, forming the core challenge addressed in this work.

\noindent\textbf{Station Graph.}~For each station-based modality $m\in\{\text{bike},\text{metro}\}$, we define a set of $N_s^m$ stations $V^m=\{v_1^m,\dots,v_{N_s^m}^m\}$. At time step $t$, each station $v_i^m$ is associated with a demand observation $d_i^{m,t}$ and a contextual feature vector that encodes location, points of interest (POIs), road-network characteristics, and weather conditions. Spatial and behavioral dependencies among stations are represented by a dynamic station graph $\mathcal{G}_s^{m,t}=(V^m,E_s^{m,t})$, where edges capture both static geographic proximity and time-varying origin--destination (OD) interaction patterns. The corresponding weighted adjacency matrix is denoted by $A_s^{m,t}$.

\noindent\textbf{Zone Graph.}~To establish a unified spatial representation across mobility modes, the city is partitioned into $N_z$ non-overlapping grid zones $\mathcal{Z}=\{z_1,z_2,\dots,z_{N_z}\}$, where each zone $z_j$ corresponds to a $1\,\text{km}\times1\,\text{km}$ spatial cell. Ride-hailing demand is naturally observed at the zone level and is denoted by $d_{z_j}^t$ at time $t$. Spatial dependencies among zones are modeled by a zone graph $\mathcal{G}_z=(\mathcal{Z},E_z)$, where edges encode geographic adjacency and road-network connectivity between neighboring zones.

\noindent\textbf{Problem Formulation.}~Given $T$ time steps of historical demand and graph observations from bike-sharing and metro systems, denoted by $\{\mathbf{d}^{m,t}, A_s^{m,t}\}_{m\in\mathcal{M}_s,\, t=1}^{T}$, together with auxiliary features, and given only static ride-hailing spatial features $\mathbf{X}^{\mathrm{RH}}_{\mathrm{spatial}}$ but no historical ride-hailing temporal observations, our objective is to forecast zone-level ride-hailing demand:
\begin{equation}
\hat{\mathbf{d}}_{z}^{\,T+1}
= f\!\left(
    \left\{\mathbf{d}^{m,t},\, A_s^{m,t}
    \right\}_{\substack{m\in\mathcal{M}_s \\ t=1,\dots,T}},\;
    \mathbf{X}^{\mathrm{RH}}_{\mathrm{spatial}}
  \right),
\end{equation}
where $\hat{\mathbf{d}}_{z}^{T+1}\in\mathbb{R}^{N_z}$ denotes the forecasted ride-hailing demand vector across all zones, and $\mathcal{M}_s=\{\text{bike},\text{metro}\}$ denotes the set of source modalities. This asymmetric availability of temporal observations between source and target systems defines the key challenge of cross-modal mobility transfer.

\begin{figure*}[t] 
  \centering 
  \includegraphics[width=\textwidth]{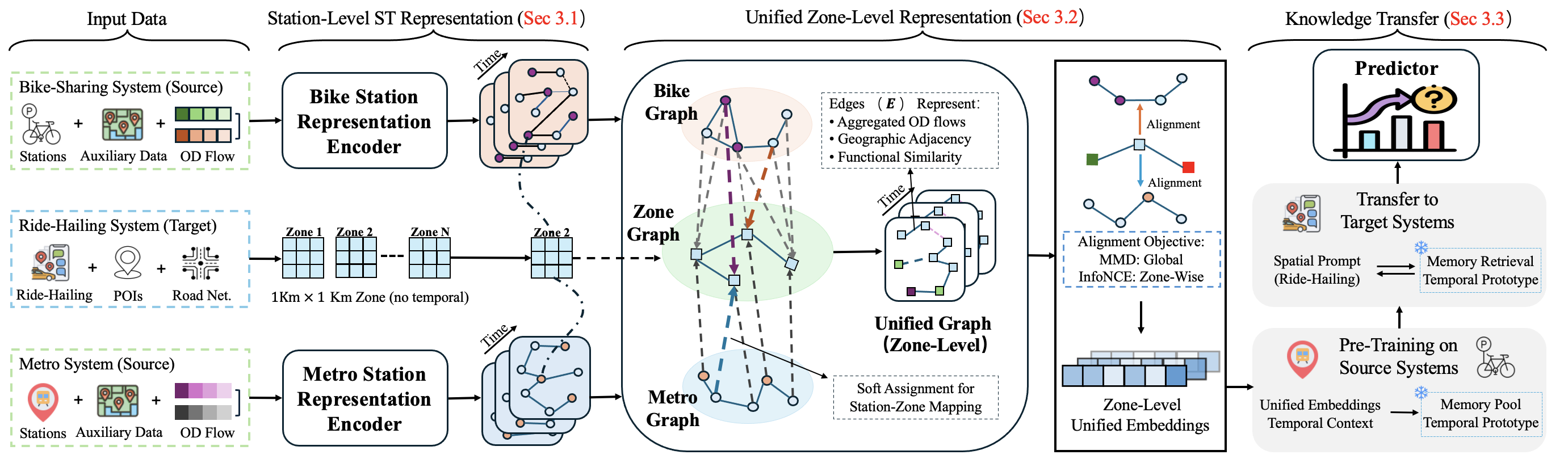} 
  \caption{Overview of the proposed TransMod framework. Station-level observations from source mobility systems are encoded to learn spatio-temporal representations, which are aggregated into a shared zone-level space through soft assignment to reconcile heterogeneous spatial granularity. Alignment across mobility modes is performed within this unified representation to reduce distributional discrepancies. Knowledge transfer across modes enables accurate demand forecasting for the target system under limited historical observations.}
  \label{fig:my_label2}  
  \Description{}
\end{figure*}

%% file: 02_Methodology.tex
\section{Methodology}

To enable demand forecasting transfer across heterogeneous mobility modes, we propose TransMod, a unified framework that integrates station-level spatio-temporal representation learning, zone-level spatial unification, and cross-modal knowledge transfer across mobility systems under sparse target observations. The overall architecture is illustrated in Figure~\ref{fig:my_label2}. The following subsections describe each component of TransMod in detail.

\subsection{Station-Level Spatio-Temporal Representation}

This module learns station-level representations by jointly modeling temporal demand dynamics and spatial interactions for bike-sharing and metro systems.

\noindent\textbf{Temporal features.}~For each source shared mobility mode, temporal representations are extracted from historical demand sequences $\mathbf{X}^m \in \mathbb{R}^{N_s^m \times T \times F}$ using dilated temporal convolutional networks to capture multi-scale patterns, yielding station embeddings $e_i^{m} \in \mathbb{R}^{d_t}$, where $m$ denotes bike-sharing or metro. To model demand uncertainty, each embedding is parameterized as a Gaussian distribution:
\begin{equation}
    \mu_i^m = \mathbf{U}^{m\top} e_i^{m}, \qquad
    \sigma_i^m = \sqrt{\text{ELU}(\mathbf{P}^{m\top} e_i^{m}) + 1},
\end{equation}
from which uncertainty-aware temporal embeddings are obtained via reparameterized sampling.

\noindent\textbf{Graph construction.}~For each mobility mode $m$, a dynamic station graph is constructed to capture spatial interactions. Each station node is initialized by concatenating its uncertainty-aware temporal embedding with spatial contextual features, including Point-of-Interest (POI) distributions $\mathbf{p}_i^m$, resulting in $\mathbf{h}_i^{m,0} = [\mu_i^m \| \mathbf{p}_i^m]$. Edges incorporate both static spatial proximity and time-varying mobility patterns. Static edges $A^m_{\text{geo}}$ connect geographically proximate stations, while dynamic edges capture temporally decayed OD flows:
\begin{equation}
    A_s^{m,t} = A^m_{\text{geo}} +
    \sum_{\tau=\max(1,t-\Delta t)}^{t}
    \exp\big(-\lambda(t-\tau)\big)\cdot F^{m,\tau},
\end{equation}
where $F^{m,\tau}$ denotes the OD flow matrix at time $\tau$. Graph attention is applied to aggregate information from neighboring stations at each time step:
\begin{equation}
    \hat{h}_i^{m,t} =
    \sigma\Big(\sum_{j \in \mathcal{N}_i^{m,t}} \alpha_{ij}^{m,t} \mathbf{W}^m h_j^{m,t}\Big),
\end{equation}
and the final station representation is $\hat{h}_i^m = \hat{h}_i^{m,T}$.

\subsection{Unified Zone-Level Representation}

To facilitate learning and transfer across heterogeneous mobility modes, we construct a unified zone-level representation that reconciles disparities in spatial granularity between station-based and zone-based systems. This unified spatial representation provides a foundation for alignment across mobility modes and subsequent urban mobility forecasting.

\subsubsection{Station-Zone Representation Aggregation}

To bridge the spatial granularity gap between station-based mobility systems (e.g., bike-sharing and metro) and zone-level modeling, we learn a soft assignment matrix for station-to-zone mapping, allowing each station to contribute to multiple zones with different weights and enabling smooth, spatially continuous aggregation.

For each station-based mobility mode $m \in \{\text{bike}, \text{metro}\}$, we learn a soft assignment matrix $S^m \in \mathbb{R}^{N_s^m \times N_z}$, where $S^m[i,j] \in [0,1]$ denotes the contribution of station $i$ to zone $j$ and satisfies the normalization constraint $\sum_j S^m[i,j] = 1$. This constraint ensures that station-level information is redistributed across zones in a normalized, stable, and interpretable manner while preserving spatial relevance. The assignment is initialized using a distance-based Gaussian kernel to encode geographical proximity:

\begin{equation}
\label{eq:soft_assign_geo}
    S^m_{\text{geo}}[i,j] =
    \frac{\exp(-d_{ij}^2 / 2\sigma_m^2)\cdot M^m[i,j]}
         {\sum_k \exp(-d_{ik}^2 / 2\sigma_m^2)\cdot M^m[i,k]},
\end{equation}
where $d_{ij}$ denotes the Euclidean distance between station $i$ and the centroid of zone $j$. The binary mask $M^m$ restricts each station to its top-$K$ nearest zones, enforcing spatial locality and avoiding unrealistic long-range associations. The mobility-specific bandwidth, $\sigma_m$, reflects differences in service coverage across mobility modes.

Starting from this geographically grounded prior, the soft assignment is refined in a data-driven manner:
\begin{equation}
    S^m = \text{Softmax}\big(\log(S^m_{\text{geo}}) + \Delta^m \odot M^m\big),
\end{equation}
where $\Delta^m$ contains learnable parameters constrained by the spatial mask. This formulation allows the assignment to adapt to observed data while remaining anchored to physically meaningful spatial relationships. To regularize the learning process and prevent excessive deviation from the geographical prior, we introduce
\begin{equation}
    \mathcal{L}_{\text{geo}} = \|S^m - S^m_{\text{geo}}\|_F^2,
\end{equation}
which encourages consistency between the learned assignment and the spatial prior.

\noindent\textbf{Zone-Level Aggregation.}~Using the learned soft assignment, station-level features and interactions are aggregated to the zone level as:
\begin{equation}
    H_z^m = (S^m)^\top \hat{H}^m, \qquad
    A_{\text{od}}^m = (S^m)^\top A_s^m S^m.
\end{equation}
For ride-hailing, which is inherently observed at the zone level, we directly construct a zone-level graph $A_z^{\text{RH}}$ by connecting geographically proximate zones within a 1\,km radius. After aggregation, all mobility modes are represented at the zone-level granularity for alignment and modeling.

\subsubsection{Unified Zone Graph Alignment}

To align heterogeneous zone-level representations from bike-sharing and metro systems within the shared latent space, we employ complementary objectives for global distribution alignment and instance-level alignment.

Global distribution alignment is achieved via Maximum Mean Discrepancy (MMD)~\cite{ghifary2014domain}:
\begin{equation}
    \mathcal{L}_{\text{MMD}} =
    \Big\|\frac{1}{N_z}\sum_i \phi(h^{\text{bike}}_{z,i})
    - \frac{1}{N_z}\sum_j \phi(h^{\text{metro}}_{z,j})\Big\|^2,
\end{equation}
where $\phi(\cdot)$ denotes a kernel-induced feature mapping.
To further enforce fine-grained alignment, an InfoNCE-based contrastive objective encourages representations from the same zone but different mobility modes to be close:
\begin{equation}
    \mathcal{L}_{\text{InfoNCE}} =
    -\sum_i \log
    \frac{\exp(\text{sim}(h^{\text{bike}}_{z,i}, h^{\text{metro}}_{z,i})/\tau)}
         {\sum_j \exp(\text{sim}(h^{\text{bike}}_{z,i}, h^{\text{metro}}_{z,j})/\tau)}.
\end{equation}

After alignment, zone-level representations from different mobility modes are projected into a shared latent space and fused to obtain unified zone embeddings $Z \in \mathbb{R}^{N_z \times d}$.

We construct a unified zone-level graph based on the unified embeddings, which captures complementary interaction patterns across modes. The adjacency matrix integrates aggregated OD flows from bike-sharing and metro systems, geographical proximity between zones, and feature similarity:
\begin{equation}
    A_z = \alpha(A^{\text{bike}}_{\text{od}} + A^{\text{metro}}_{\text{od}})
        + \beta A_{\text{geo}} + \gamma A_{\text{feat}},
\end{equation}
where the coefficients $\alpha$, $\beta$, and $\gamma$ control the contributions of mobility interactions, spatial adjacency, and semantic similarity, respectively. The resulting graph is sparsified to retain the most informative zone connections, yielding a compact zone graph for subsequent representation learning.

\subsection{Knowledge Transfer Across Modes}

Building on the station-to-zone spatial alignment, we introduce a memory-based mechanism to transfer temporal knowledge from data-rich bike-sharing and metro systems to ride-hailing (RH), where historical temporal observations are limited or unavailable.

We maintain a learnable memory pool $\mathcal{M}=\{(\mathbf{k}_i,\mathbf{v}_i)\}_{i=1}^{M}$ with $M$ key--value pairs, where $\mathbf{k}_i,\mathbf{v}_i \in \mathbb{R}^{d}$ encode transferable temporal prototypes. Given a query vector $\mathbf{q}_j$ for zone $z_j$, relevant temporal priors are retrieved through an attention-based memory lookup:
\begin{equation}
\label{eq:memory_lookup}
\mathbf{m}_j = \sum_{i=1}^{M} \alpha_{j,i}\mathbf{v}_i, \quad
\alpha_{j,i}=\text{softmax}_i\Big(\frac{\mathbf{k}_i^\top
\mathbf{q}_j}{\sqrt{d}}\Big).
\end{equation}

\noindent\textbf{Pre-training on source systems.}~For each zone, the query is constructed by fusing aligned zone-level representations from bike-sharing and metro systems with temporal context features, such as hour of the day, day of the week, and holidays:
\begin{equation}
\mathbf{q}_j = \text{MLP}\!\left(
[H^{\text{bike}}_{z,j} \;\|\; H^{\text{metro}}_{z,j} \;\|\;
\mathbf{t}]\right).
\end{equation}
Since each query jointly encodes the zone-specific spatial context and the current temporal state, the memory keys $\mathbf{k}_i$ are optimized to index patterns at the intersection of \emph{where} the demand occurs and \emph{when} it emerges. Therefore, the memory does not simply function as a time-of-day lookup table; instead, it learns reusable spatio-temporal prototypes that capture mode-shared demand regularities. The diversity regularizer $\mathcal{L}_{\text{div}}$ further encourages the $M$ prototypes to specialize in distinct spatio-temporal configurations, preventing them from collapsing into a shared average representation. The retrieved memory vector $\mathbf{m}_j$ is then fused with zone representations and propagated over the zone graph for demand prediction.

\noindent\textbf{Transfer to target system.}~During transfer, the memory pool and spatial mapping modules are frozen to preserve the transferable patterns learned from source systems. Since RH lacks historical temporal observations, query prompts are generated solely from RH spatial cues through a lightweight prompt network, where the prompt for zone $z_j$ is defined as $\mathbf{p}^{\text{RH}}_j = \text{MLP}(\mathbf{s}_j)$. Here, $\mathbf{s}_j$ aggregates zone-level POI distributions, road-network characteristics, and graph structural indicators for zone $z_j$. Temporal priors are then retrieved through the same attention-based memory lookup defined in Eq.~(\ref{eq:memory_lookup}), with $\mathbf{p}^{\text{RH}}_j$ used in place of $\mathbf{q}_j$. Since the memory keys are trained during source pre-training to associate spatial contexts with temporal regimes, the RH spatial prompt can retrieve zone-appropriate temporal priors from the frozen memory. This realizes a spatial-to-temporal mapping without requiring target-domain temporal history. During adaptation, only the prompt network and prediction head are fine-tuned, enabling knowledge transfer while retaining the temporal knowledge encoded in the memory pool.

\subsection{Training Objective}

During \textbf{pre-training}, we jointly optimize demand prediction and representation learning across multiple source mobility modes using a composite objective:
\begin{equation}
\mathcal{L}_{\text{pre}} =
\mathcal{L}_{\text{pred}}
+ \lambda_{\text{ali}} \mathcal{L}_{\text{ali}}
+ \lambda_{\text{geo}} \mathcal{L}_{\text{geo}}
+ \lambda_{\text{div}} \mathcal{L}_{\text{div}},
\end{equation}
\noindent where $\mathcal{L}_{\text{pred}}$ denotes the forecasting loss on bike-sharing and metro systems. $\mathcal{L}_{\text{ali}}$ enforces zone-level alignment across modes via MMD and InfoNCE. $\mathcal{L}_{\text{geo}}$ regularizes soft assignments toward geographically constrained initialization. $\mathcal{L}_{\text{div}}$ encourages diversity among memory prototypes to prevent representation collapse. The hyperparameters $\lambda_{\text{align}}$, $\lambda_{\text{geo}}$, and $\lambda_{\text{div}}$ balance the contributions of individual components.

During ride-hailing \textbf{adaptation}, the memory pool $\mathcal{M}$ and spatial mapping modules are frozen to preserve the transferable spatio-temporal knowledge learned from source systems. Only the prompt network and prediction head are fine-tuned using the ride-hailing demand prediction loss:
\begin{equation}
\mathcal{L}_{\text{t}} =
\text{MSE}(\hat{d}^{\text{RH}}_z, d^{\text{RH}}_z).
\end{equation}
This training strategy limits trainable parameters and preserves the temporal priors encoded in the memory.

%% file: 03_Experiments.tex
\section{Experiments}

In this section, we present the implementation details and dataset descriptions, followed by the evaluation metrics and baseline methods used for comparison. We then conduct extensive experiments to assess TransMod's performance, including comparisons with baseline models, computing efficiency analysis, ablation studies, and robustness analysis across different scenarios.

\subsection{Implementation Details}

We set the hidden dimension to $d=64$ and employ 2-layer GATs with 4 attention heads for both station and zone-level graphs. Soft assignment is initialized using Gaussian kernels with $\sigma_{\text{bike}}=500$m and $\sigma_{\text{metro}}=800$m, retaining the top-$K$ nearest zones per station. The memory pool contains $M=128$ prototypes. Models are trained using Adam with learning rate $10^{-3}$ and batch size 32. The loss weights are set to $\lambda_{\text{MMD}}=0.1$, $\lambda_{\text{InfoNCE}}=0.5$, $\lambda_{\text{geo}}=0.1$, and $\lambda_{\text{div}}=0.05$, with InfoNCE temperature $\tau=0.07$ and diversity threshold $\delta=0.3$. Ride-hailing adaptation is conducted for 50 epochs with learning rate $5\times10^{-4}$. Demand is aggregated hourly with temporal window $T=12$, and data are split into 70\%/15\%/15\% for training, validation, and testing. All experiments are conducted on an NVIDIA RTX 4090 GPU. The model implementation is available at: \url{https://anonymous.4open.science/r/TransMod-DDA5}.

\subsection{Data Description}
  
To evaluate our framework, we collect multiple urban shared mobility datasets from New York City\footnote{https://opendata.cityofnewyork.us/} and Chicago\footnote{https://data.cityofchicago.org/}, integrating bike-sharing, metro, and ride-hailing records with rich auxiliary features, including points of interest (POI), meteorological conditions, and road network information.

\noindent\textbf{Bike-Sharing Data.}~We use operational records from NYC Citi Bike and Chicago Divvy for January 2018. The data contain station-level pick-up and drop-off events with timestamps and precise geographic coordinates.

\noindent\textbf{Metro Data.}~Metro ridership data are obtained from the NYC MTA and the Chicago CTA, covering 472 stations in NYC and 145 stations in Chicago. These datasets provide hourly entry and exit counts along with station metadata.

\noindent\textbf{Ride-Hailing Data.}~Ride-hailing trip records are collected from the NYC TLC and City of Chicago TNP Trips dataset. Trips are aggregated into predefined spatial zones, with pick-up and drop-off timestamps and locations.

\begin{table}[t]
\centering
\caption{Stage-wise data access protocol.
  \checkmark~= used; $\times$~= unavailable;
  \textsc{frz}~= frozen, not re-accessed.}
\label{tab:mytable3}
\footnotesize
\setlength{\tabcolsep}{3pt}
\begin{tabularx}{\linewidth}{lXccc}
\toprule
\textbf{Input} & \textbf{Type} &
  \textbf{Pre.} & \textbf{Trans.} & \textbf{Eval.} \\
\midrule
Bike/metro demand
  & Hourly sequences
  & \checkmark & \textsc{frz} & \textsc{frz} \\
Bike/metro OD
  & Temporal flow matrices
  & \checkmark & \textsc{frz} & \textsc{frz} \\
Auxiliary feat.
  & POI, weather, road
  & \checkmark & \textsc{frz} & \textsc{frz} \\
\midrule
RH hist. demand
  & Hourly sequences
  & $\times$ & $\times$ & $\times$ \\
RH spatial feat.
  & POI, road, geometry
  & $\times$ & \checkmark & \checkmark \\
RH test demand
  & Held-out labels
  & $\times$ & $\times$ & \checkmark \\
\bottomrule
\end{tabularx}
\end{table}

\noindent\textbf{Auxiliary Data.}~To enrich spatial and contextual representations, we incorporate POI data from the Google Places API, meteorological data from MesoWest, and road network data from OpenStreetMap.

\noindent\textbf{Data Protocol.}~To prevent target-domain leakage, TransMod enforces a strict stage-wise data separation. Ride-hailing temporal demand is not used during source pre-training or as historical temporal input during transfer. During target adaptation, only limited ride-hailing labels are used for supervised fine-tuning, while held-out ride-hailing records are reserved exclusively for evaluation. Table~\ref{tab:mytable3} summarizes the input availability at each stage.

\begin{table*}[t]
  \centering
  \caption{Overall Performance Comparison on Transfer Tasks across Mobility Modes.~The best results are marked in bold.}
  \resizebox{\textwidth}{!}{
  \begin{tabular}{lccc ccc}
    \toprule
    \multirow{3}{*}{Model} 
    & \multicolumn{3}{c}{ \textbf{New York City}}
    & \multicolumn{3}{c}{\textbf{Chicago City}} \\
    \cmidrule(lr){2-4}\cmidrule(lr){5-7}
    & MAE $\downarrow$ & RMSE $\downarrow$ & MAPE(\%) $\downarrow$
    & MAE $\downarrow$ & RMSE $\downarrow$ & MAPE(\%) $\downarrow$ \\
    \midrule
    ARIMA            & 18.92$\pm$1.25 & 23.35$\pm$1.56 & 16.61$\pm$1.08 & 15.48$\pm$1.02 & 19.90$\pm$1.44 & 17.82$\pm$1.28 \\
    LSTM             & 17.63$\pm$1.84 & 22.97$\pm$2.07 & 15.95$\pm$1.25 & 14.36$\pm$0.31 & 18.27$\pm$0.49 & 16.54$\pm$1.15 \\
    \midrule
    RegionTrans      & 13.78$\pm$1.24 & 18.92$\pm$1.70 & 11.82$\pm$1.38 & 12.53$\pm$1.02 & 16.21$\pm$1.44 & 10.34$\pm$1.19 \\
    STAN             & 13.94$\pm$2.75 & 17.63$\pm$1.52 & 11.23$\pm$1.25 & 11.82$\pm$0.89 & 15.28$\pm$1.31 & 9.74$\pm$1.08 \\
    TSJT             & 13.52$\pm$1.14 & 17.15$\pm$1.48 & 10.94$\pm$3.31 & 11.45$\pm$0.92 & 14.82$\pm$2.76 & 9.42$\pm$1.12 \\
    HimNet           & 13.08$\pm$3.02 & 16.54$\pm$1.41 & 10.61$\pm$1.18 & 11.02$\pm$0.85 & 14.35$\pm$1.19 & 9.15$\pm$1.02 \\
    TransGTR         & 12.42$\pm$0.96 & 15.68$\pm$1.35 & 10.12$\pm$1.09 & 10.51$\pm$0.78 & 13.64$\pm$5.22 & 8.73$\pm$0.95 \\
    MetaST           & 12.85$\pm$1.05 & 16.21$\pm$3.38 & 10.45$\pm$1.15 & 10.83$\pm$0.83 & 14.01$\pm$3.15 & 9.02$\pm$1.04 \\
    \midrule
    LSTM-FT          & 13.68$\pm$1.12 & 17.34$\pm$1.46 & 10.98$\pm$1.22 & 11.58$\pm$0.91 & 15.01$\pm$1.24 & 9.51$\pm$1.10 \\
    MMDNet           & 12.94$\pm$1.08 & 16.38$\pm$1.42 & 10.52$\pm$1.18 & 10.92$\pm$0.86 & 14.18$\pm$1.18 & 9.08$\pm$1.05 \\
    FusionTN         & 12.61$\pm$0.98 & 15.93$\pm$1.36 & 10.24$\pm$1.11 & 10.64$\pm$0.81 & 13.82$\pm$1.14 & 8.84$\pm$0.98 \\
\midrule
\textbf{TransMod (Ours)}
  & \textbf{10.78$\pm$0.25} & \textbf{13.16$\pm$0.73} & \textbf{8.76$\pm$0.85}
  & \textbf{8.30$\pm$0.16} & \textbf{11.37$\pm$0.64} & \textbf{7.52$\pm$0.78} \\
    \bottomrule
  \end{tabular}}
  \label{tab:my_label1}
\end{table*} 

\subsection{Evaluation Metrics and Compared Methods}

\textbf{Evaluation metrics}.~There are three performance metrics employed for evaluating our model, namely Root Mean Square Error (RMSE), Mean Absolute Error (MAE), and Mean Absolute Percentage Error (MAPE). These metrics collectively provide a robust and complementary framework for assessing our model's ability to capture subtle variations and forecasting errors in the data.

\noindent\textbf{Compared methods.}~We compare TransMod with baseline methods from three categories: single-system forecasting, spatio-temporal demand modeling, and transfer/cross-modal learning.

\begin{itemize}[leftmargin=*, itemsep=2pt, topsep=3pt]
    \item \textbf{Single-system models}: ARIMA~\cite{10.5555/561899}, a classical statistical time-series forecasting method, and LSTM~\cite{hochreiter1997long}, a recurrent neural network widely used for sequence prediction.

    \item \textbf{Spatio-temporal models}: RegionTrans~\cite{10.5555/3367243.3367301}, STAN~\cite{ijcai2022p282}, TSJT~\cite{Zhang_Wang_Yu_Sun_Wang_Wang_2025}, and HimNet~\cite{10.1145/3637528.3671961}, which capture spatial dependencies, temporal dynamics, and urban heterogeneity for demand forecasting, without explicitly modeling cross-modal transfer.

    \item \textbf{Transfer and cross-modal models}: TransGTR~\cite{jin2023transferable}, MetaST~\cite{yao2019learning}, LSTM-FT~\cite{hua2025transfer}, MMDNet~\cite{jiang2025leveraging}, and FusionTransNet (FusionTN)~\cite{wang2024fusiontransnet}, which leverage transferable knowledge or shared representations across regions, systems, or mobility modes.
\end{itemize}

\begin{figure*}[tb]
\centering
\begin{minipage}[b]{.23\textwidth}
    \centering
    \includegraphics[width=\linewidth]{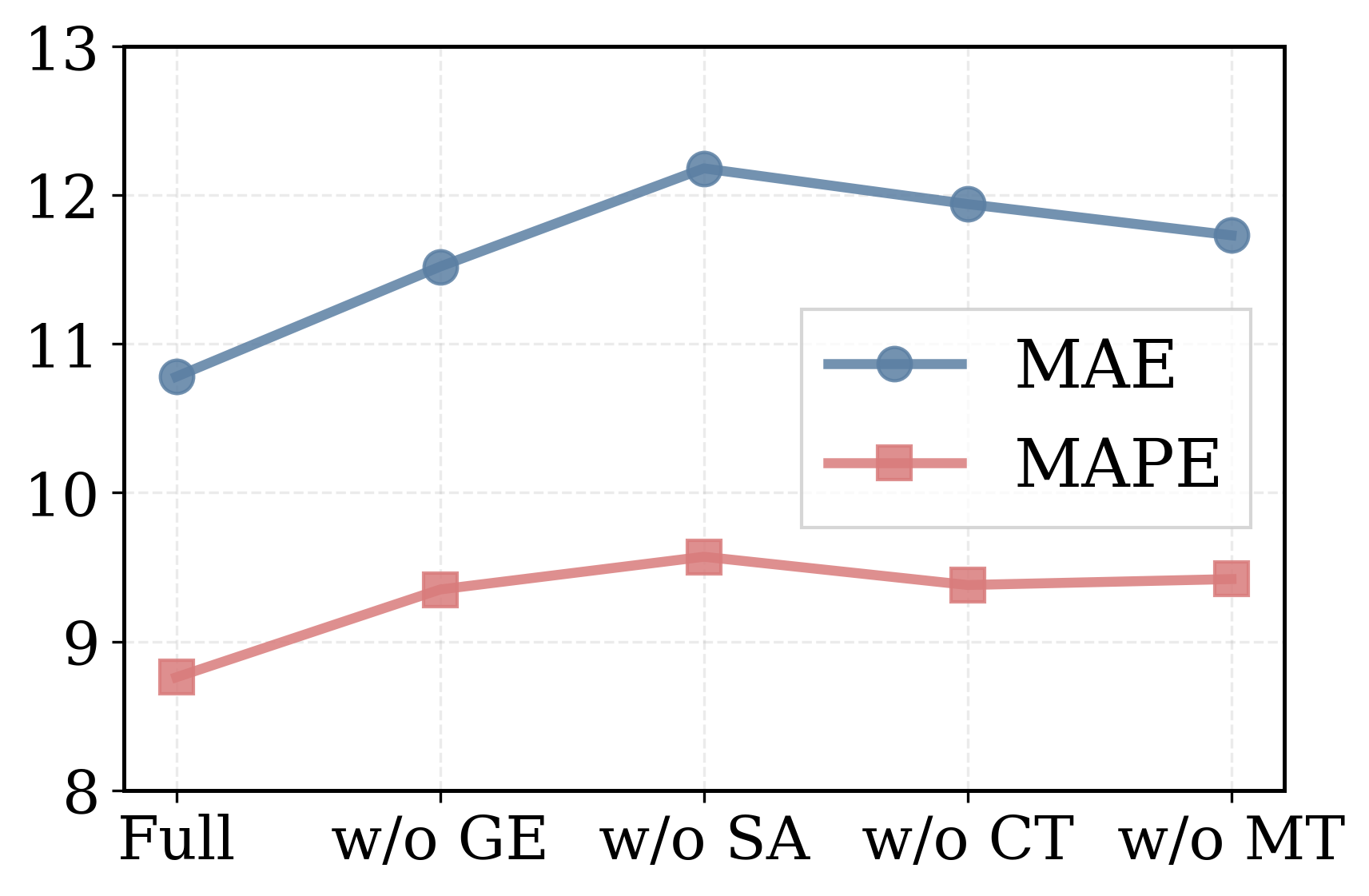}
    \subcaption{}\label{fig:subfig_a3}
\end{minipage}\hfill
\begin{minipage}[b]{.23\textwidth}
    \centering
    \includegraphics[width=\linewidth]{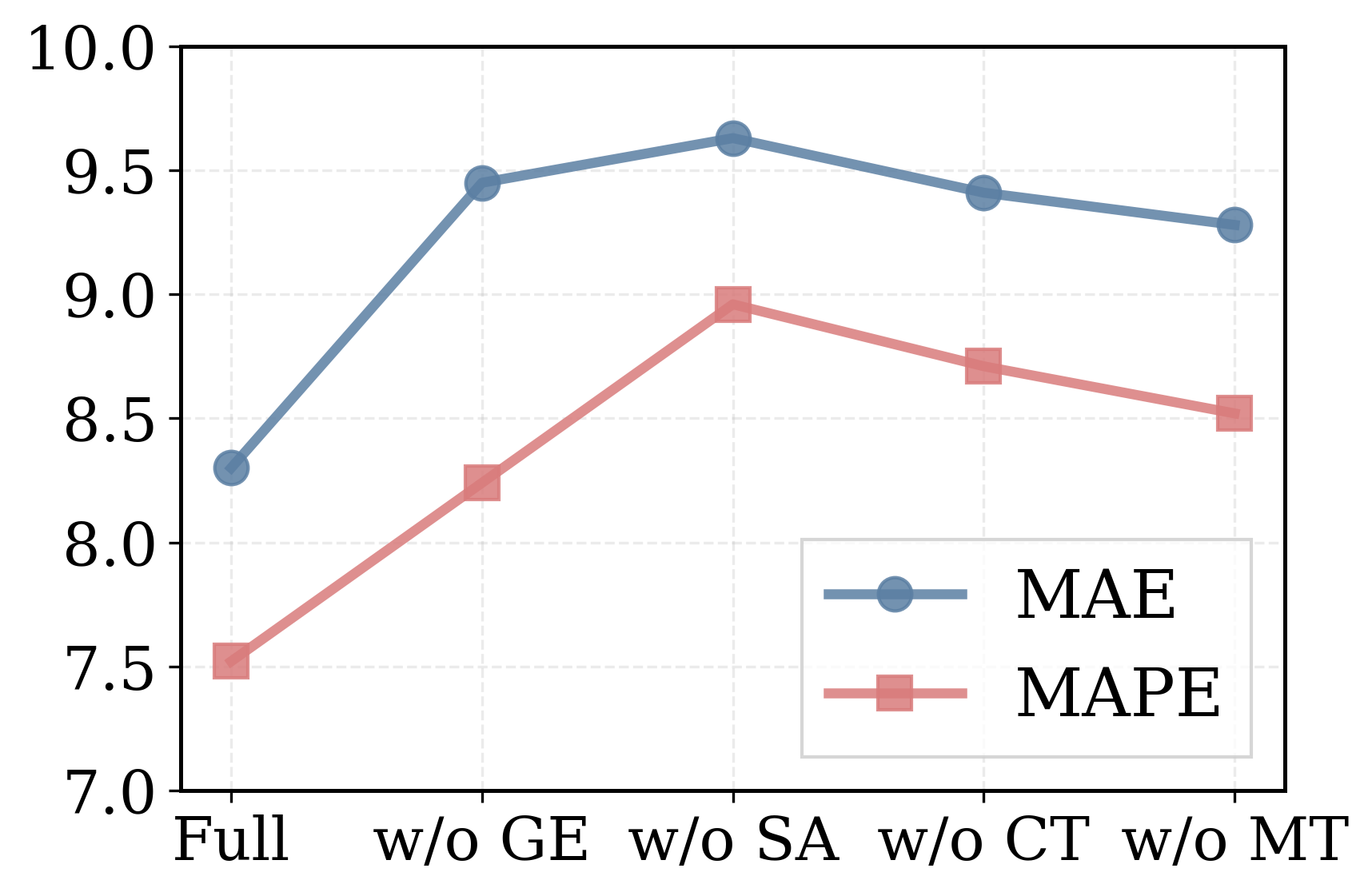}
    \subcaption{}\label{fig:subfig_b3}
\end{minipage}\hfill
\begin{minipage}[b]{.23\textwidth}
    \centering
    \includegraphics[width=\linewidth]{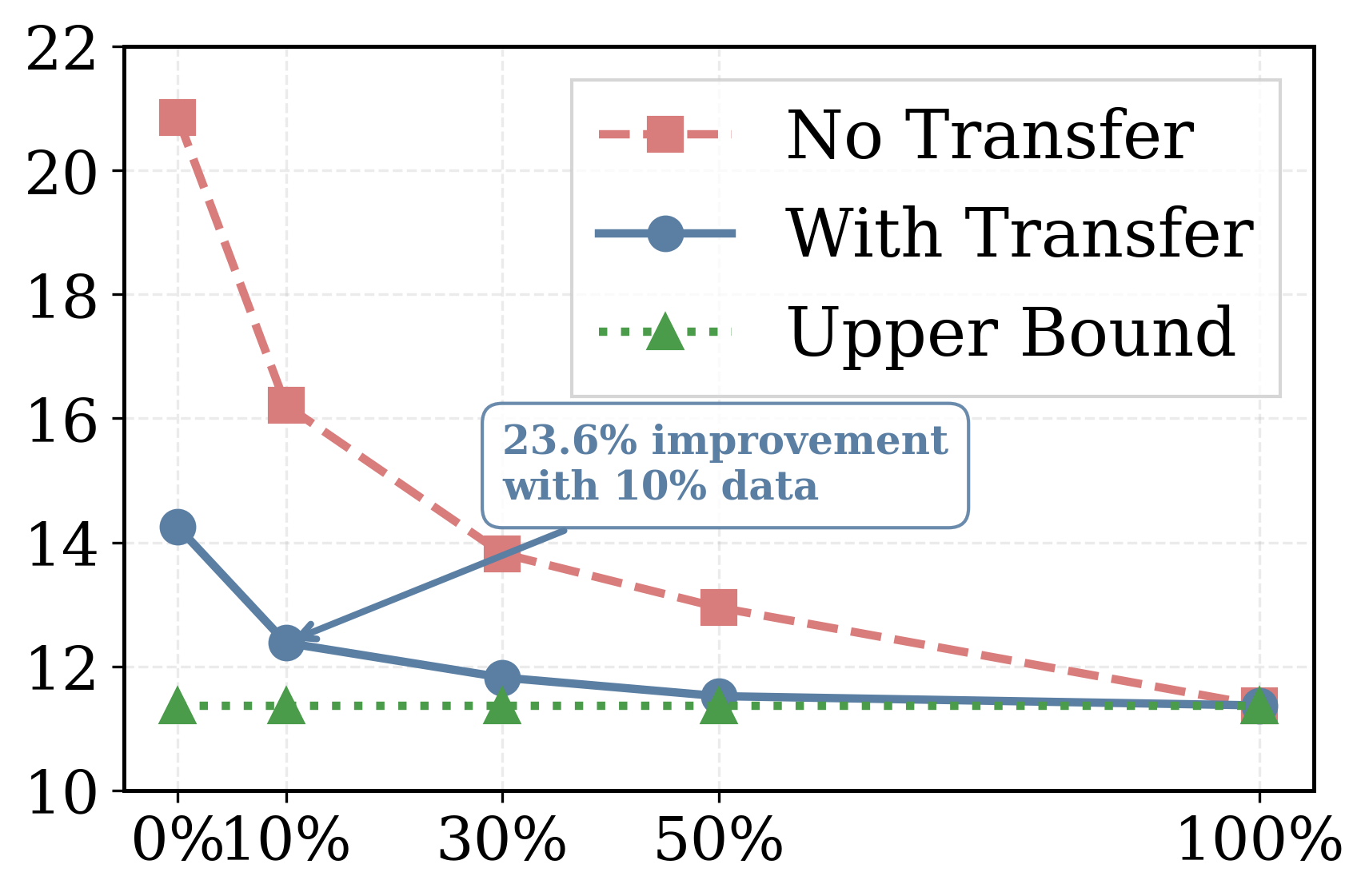}
    \subcaption{}\label{fig:subfig_a4}
\end{minipage}\hfill
\begin{minipage}[b]{.23\textwidth}
    \centering
    \includegraphics[width=\linewidth]{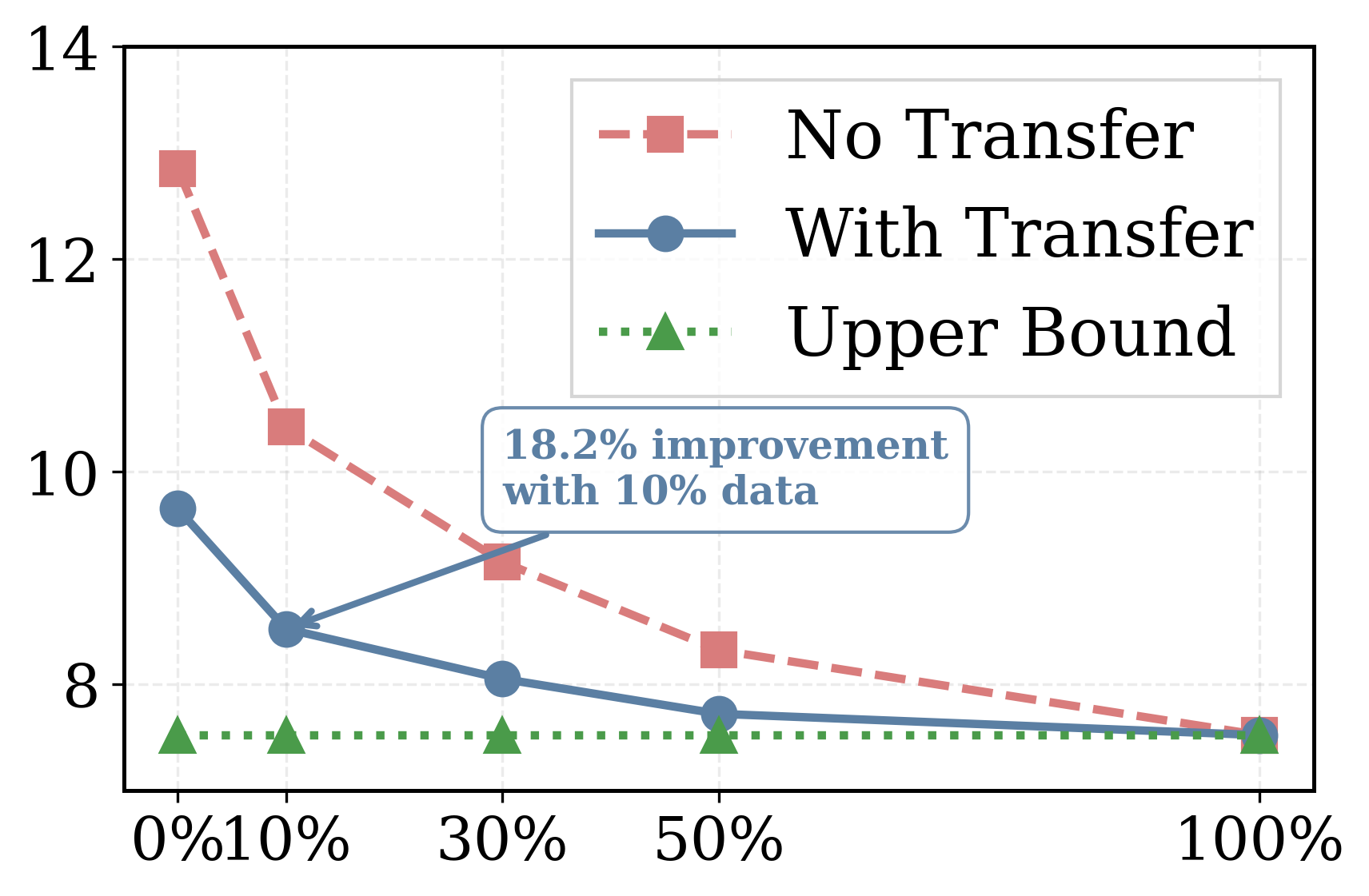}
    \subcaption{}\label{fig:subfig_b4}
\end{minipage}
\begin{minipage}[b]{.23\textwidth}
    \centering
    \includegraphics[width=\linewidth]{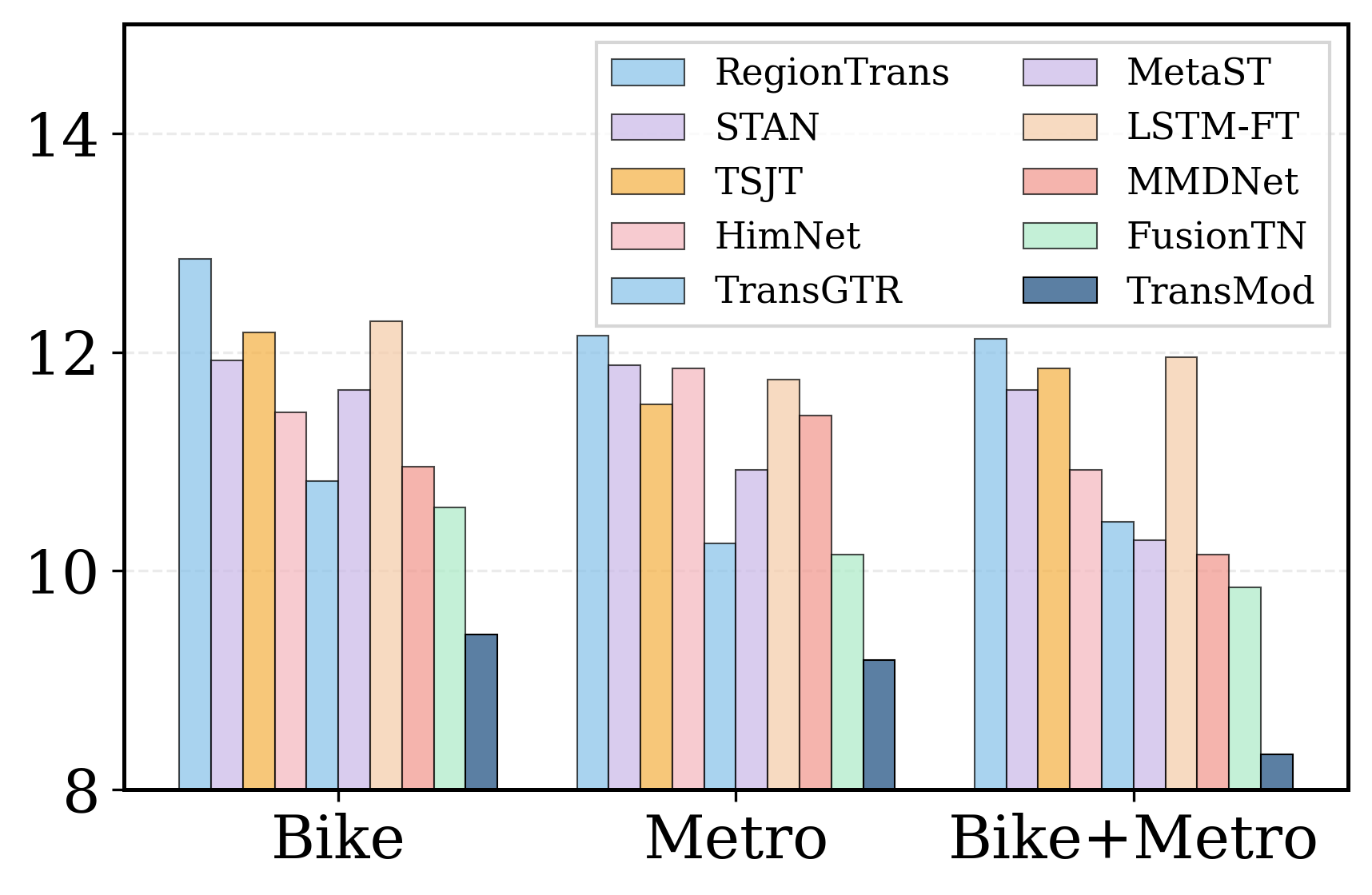}
    \subcaption{}\label{fig:subfig_a5}
\end{minipage}\hfill
\begin{minipage}[b]{.23\textwidth}
    \centering
    \includegraphics[width=\linewidth]{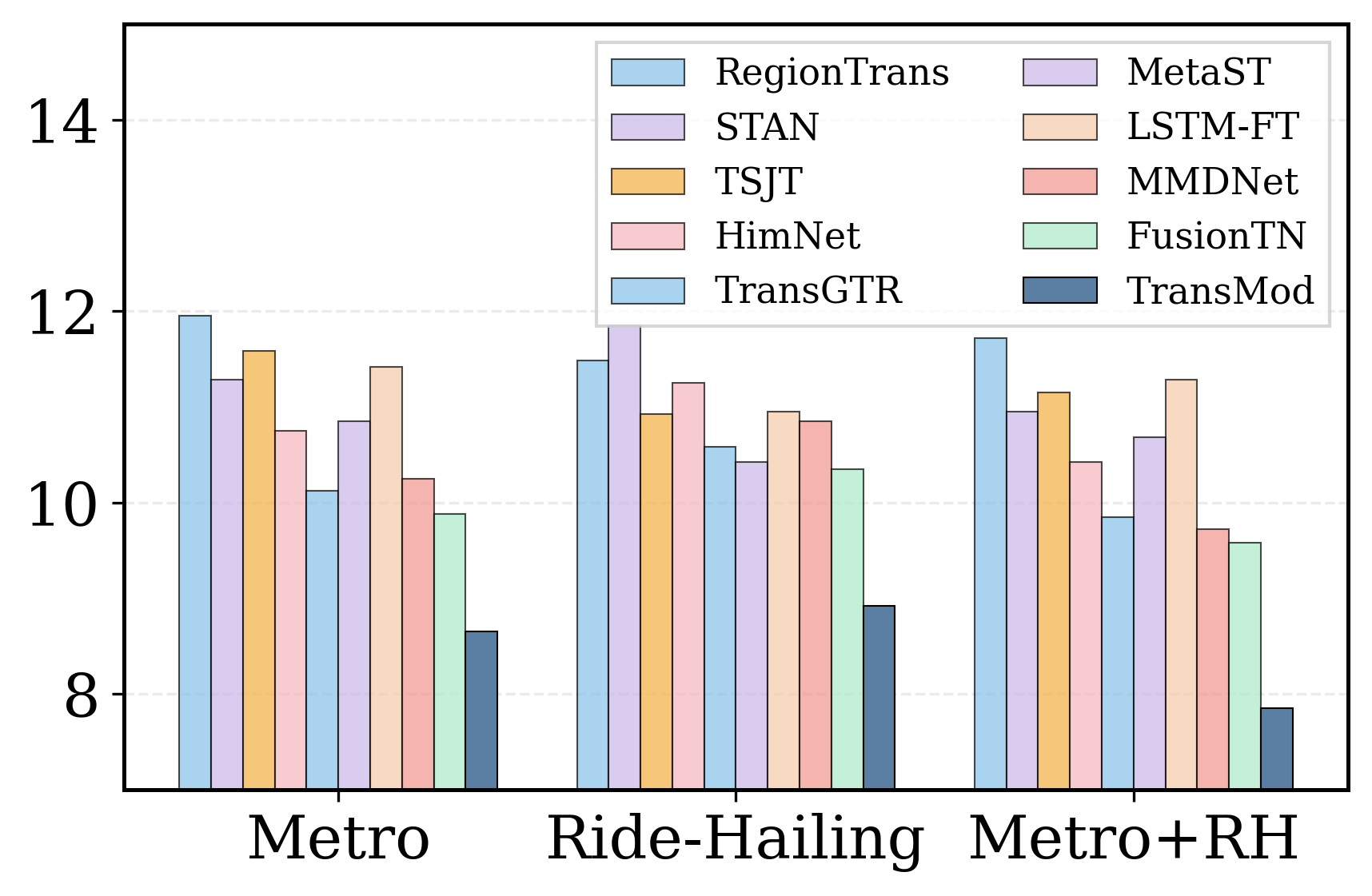}
    \subcaption{}\label{fig:subfig_b5}
\end{minipage}\hfill
\begin{minipage}[b]{.23\textwidth}
    \centering
    \includegraphics[width=\linewidth]{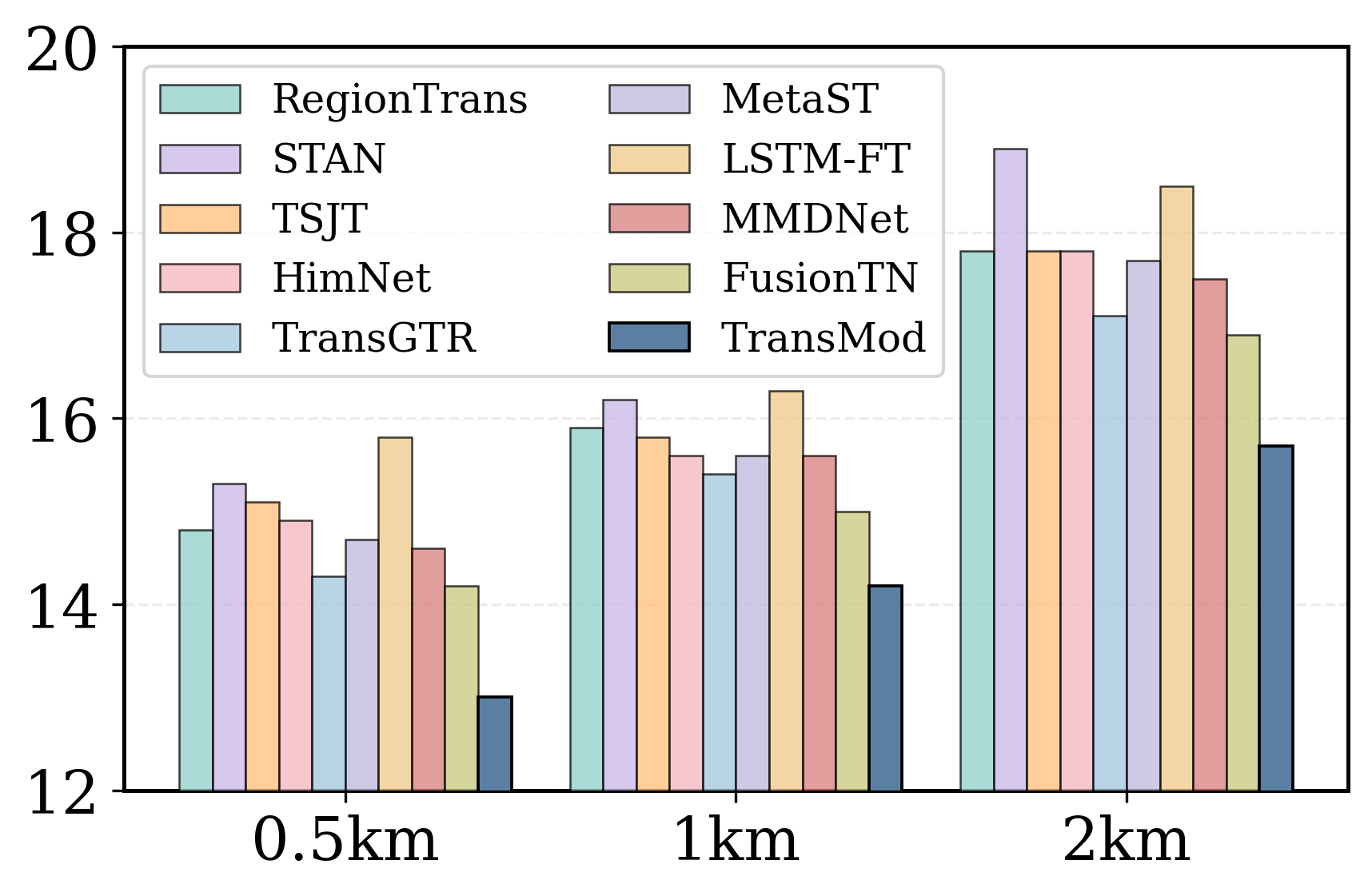}
    \subcaption{}\label{fig:subfig_a6}
\end{minipage}\hfill
\begin{minipage}[b]{.23\textwidth}
    \centering
    \includegraphics[width=\linewidth]{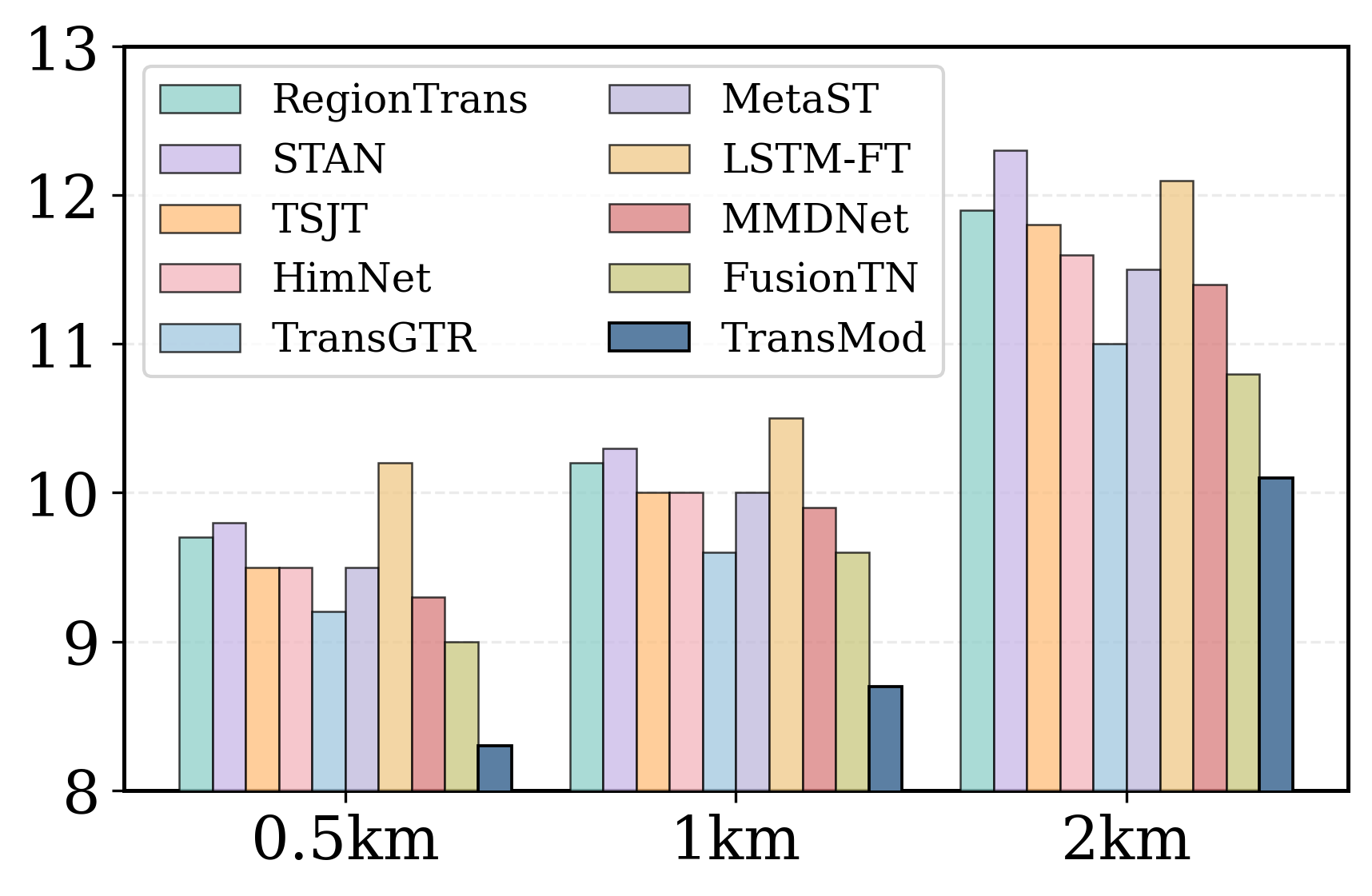}
    \subcaption{}\label{fig:subfig_b6}
\end{minipage}
\begin{minipage}[b]{.23\textwidth}
    \centering
    \includegraphics[width=\linewidth]{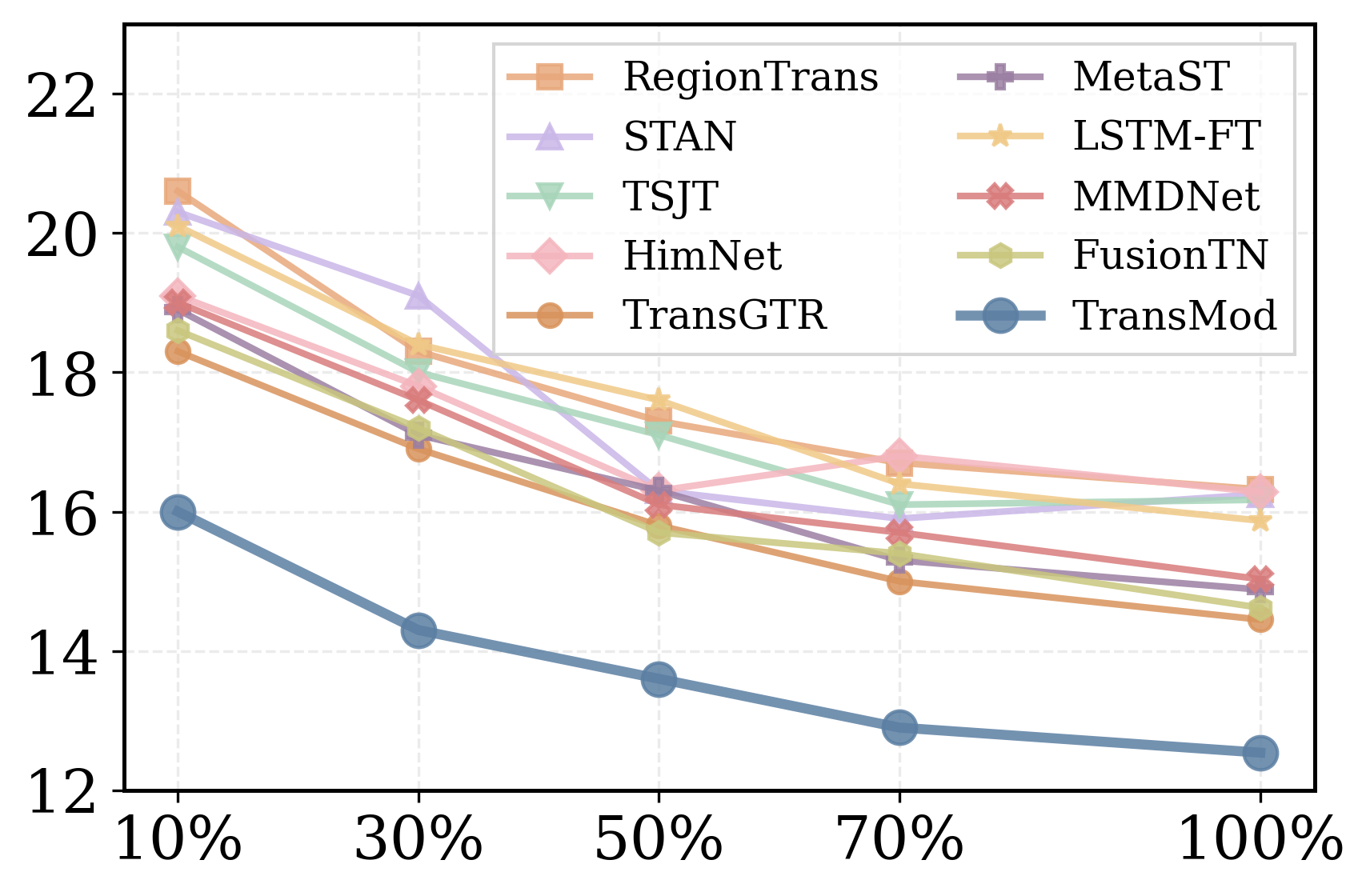}
    \subcaption{}\label{fig:subfig_a7}
\end{minipage}\hfill
\begin{minipage}[b]{.23\textwidth}
    \centering
    \includegraphics[width=\linewidth]{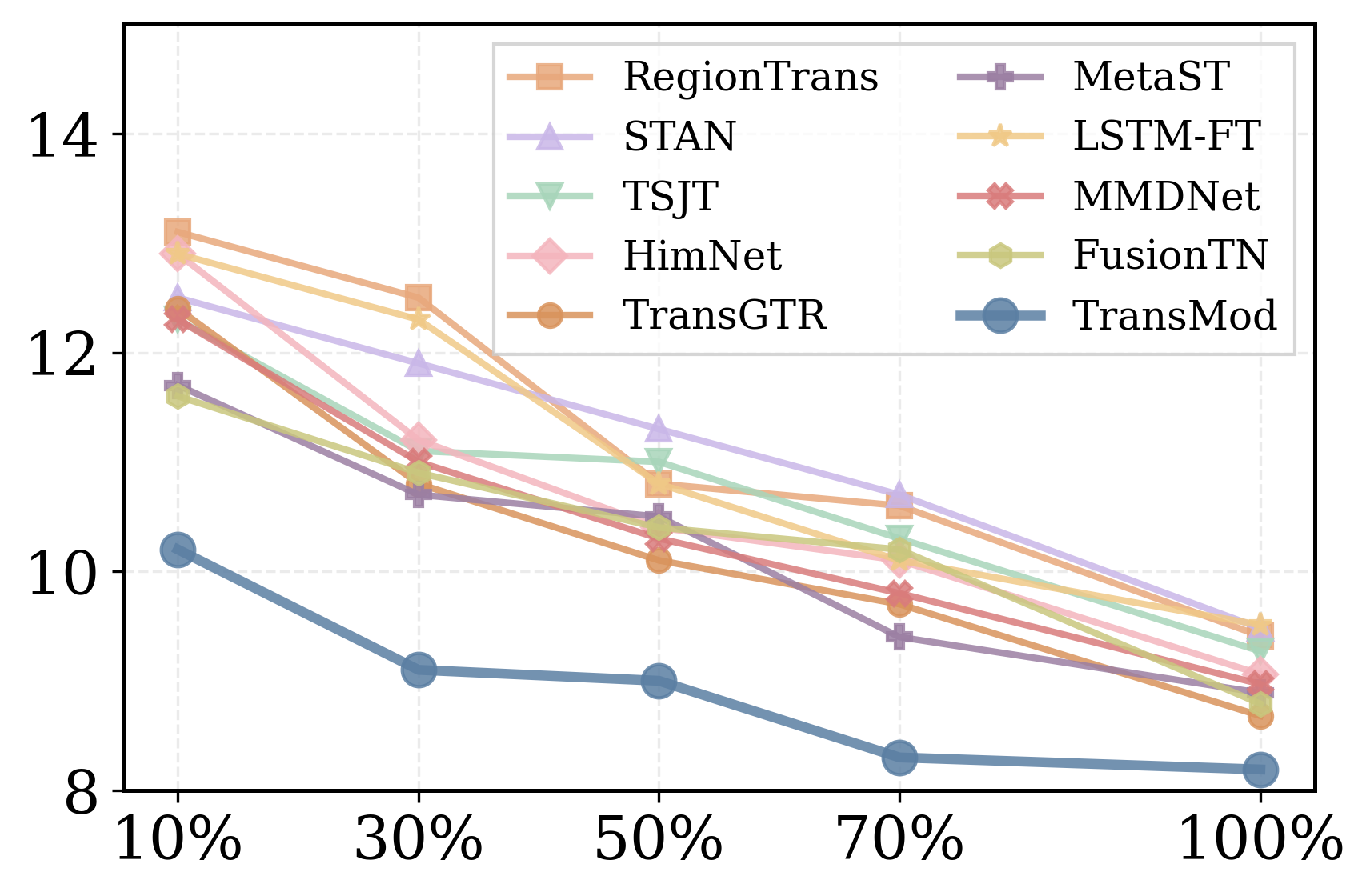}
    \subcaption{}\label{fig:subfig_b7}
\end{minipage}\hfill
\begin{minipage}[b]{.23\textwidth}
    \centering
    \includegraphics[width=\linewidth]{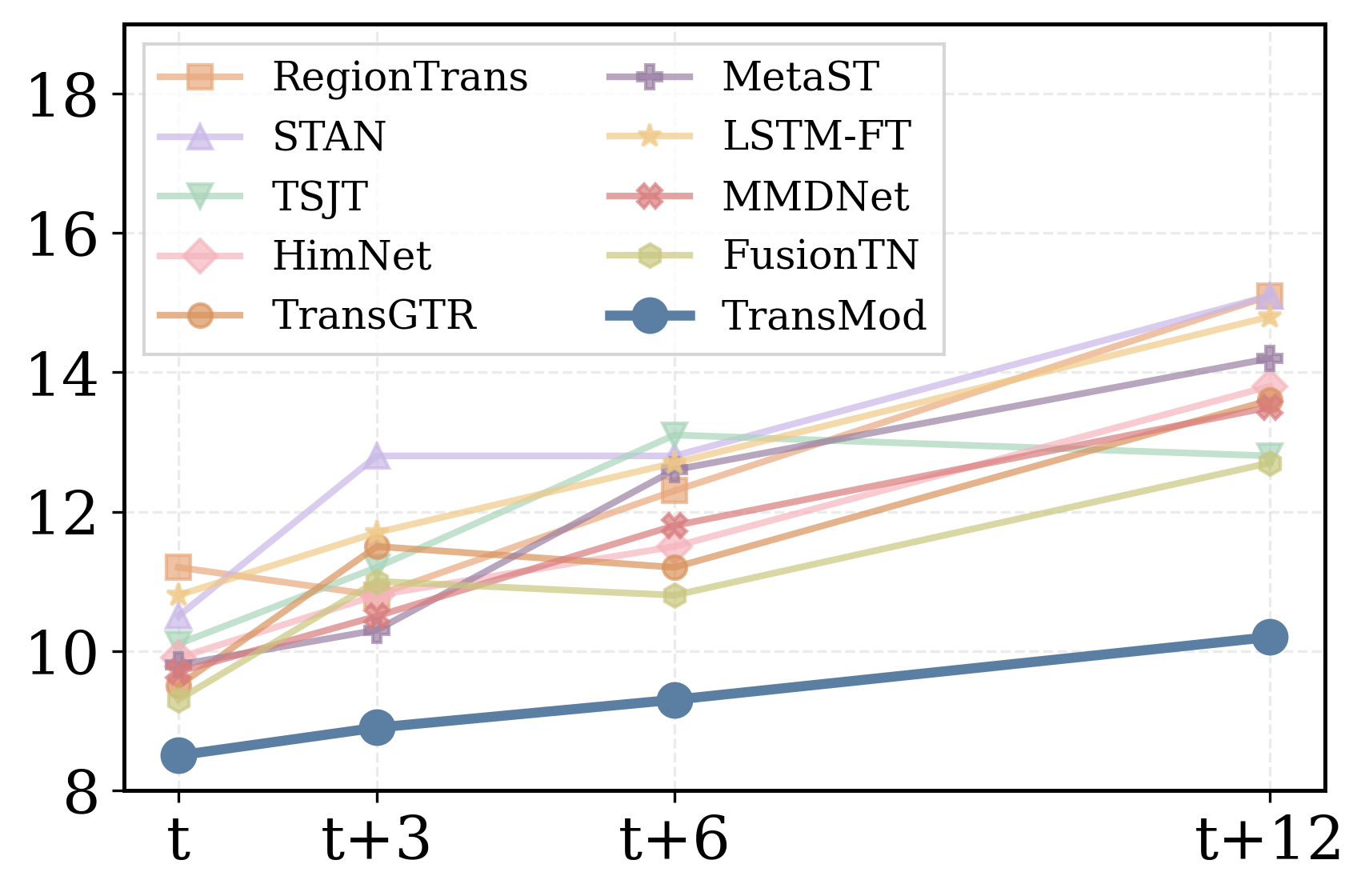}
    \subcaption{}\label{fig:subfig_a8}
\end{minipage}\hfill
\begin{minipage}[b]{.23\textwidth}
    \centering
    \includegraphics[width=\linewidth]{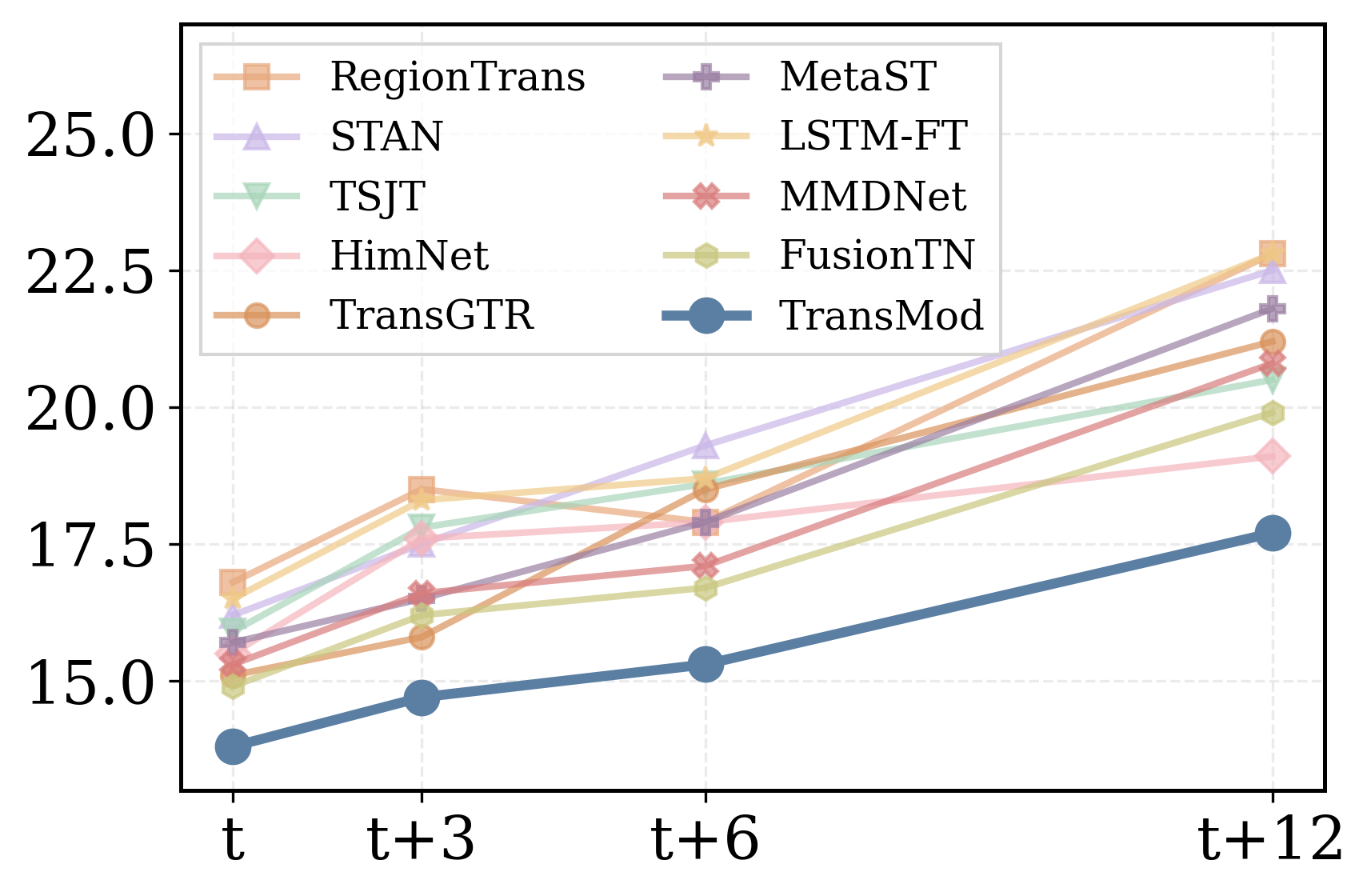}
    \subcaption{}\label{fig:subfig_b8}
\end{minipage}
\caption{Ablation studies on the (a) NYC and (b) Chicago, showing MAE and MAPE for different variants. Prediction performance for cross-city transfer from NYC to Chicago in terms of (c) RMSE and (d) MAPE. Cross-modal transfer robustness on the (e) ride-hailing and (f) bike-sharing demand forecasting tasks, shown as MAPE across different source modality configurations. Performance under different spatial resolutions, evaluated using (g) RMSE and (h) MAPE. Results under different levels of data scarcity measured by (i) RMSE and (j) MAPE. Temporal prediction performance under different prediction lengths, reported by (k) RMSE and (l) MAPE.}
\Description{}
\label{fig:my_label3}
\end{figure*}

\subsection{Comparison with Baselines}

A comprehensive comparison of demand forecasting models for transfer from station-based systems to zone-level ride-hailing demand in New York City and Chicago is reported in Table~\ref{tab:my_label1}. Traditional single-system baselines, including ARIMA and LSTM, yield substantially higher MAE and RMSE values, reflecting their limited ability to capture complex urban mobility dynamics when transferable information across systems is unavailable. Region-level and spatio-temporal models without explicit transfer across mobility modes, such as RegionTrans, STAN, TSJT, and HimNet, achieve improved performance by exploiting spatial correlations and regional structures. Nevertheless, their predictive accuracy remains constrained by the fundamental mismatch between station-level observations and zone-level prediction targets. Transfer learning approaches, including TransGTR, MetaST, and LSTM-FT, together with recent cross-modal mobility models such as MMDNet and FusionTransNet, further improve performance by leveraging shared representations across systems. However, these methods still struggle to fully resolve spatial granularity discrepancies and temporal sparsity in the target system. 

\begin{table}[t]
\centering
\caption{Computational efficiency comparison on NYC.
Times are reported in s/ep for pre-training/transfer and ms/sample for inference latency.}
\label{tab:my_label2}
\footnotesize
\setlength{\tabcolsep}{4pt}
\begin{tabularx}{\linewidth}{lCCC}
\toprule
\textbf{Method} & \textbf{Pre-train} & \textbf{Trans.} & \textbf{Infer.} \\
 & \textbf{(s/ep)} & \textbf{(s/ep)} & \textbf{(ms)} \\
\midrule
LSTM-FT    & 16.4 & 13.7 &  3.2 \\
TransGTR   & 26.5 & 24.3 & 12.8 \\
MetaST     & 31.2 & 29.8 & 14.3 \\
MMDNet     & 37.8 & 32.5 & 15.6 \\
FusionTN   & 43.8 & 41.2 & 18.4 \\
\midrule
\textbf{TransMod} & 38.4 & \textbf{5.2} & \textbf{9.8} \\
\bottomrule
\end{tabularx}
\end{table}

The proposed TransMod framework consistently outperforms all competing methods across all evaluation metrics and cities. As shown in Table~\ref{tab:my_label1}, TransMod achieves the lowest MAE, RMSE, and MAPE, demonstrating its superior generalization ability across heterogeneous mobility systems under diverse real-world settings. This performance advantage stems from its unified treatment of heterogeneity across mobility modes, which jointly addresses station–zone aggregation through geography-constrained soft assignment and temporal knowledge transfer through memory prompting. By aligning heterogeneous representations and compensating for missing temporal signals in ride-hailing demand, TransMod provides a robust and accurate solution for urban demand forecasting across mobility systems.

\subsection{Computing Efficiency Analysis}

We evaluate the computational efficiency of TransMod against all baselines on the NYC dataset using an NVIDIA RTX 4090 GPU. Table~\ref{tab:my_label2} reports three efficiency indicators: per-epoch pre-training time, per-epoch adaptation time, and per-sample inference latency. 

TransMod requires a pre-training cost comparable to other graph-based transfer methods, mainly due to cross-modal representation alignment and memory learning. However, its adaptation stage is substantially more efficient than competing transfer methods, requiring only 5.2\,s/ep compared with 24.3--41.2\,s/ep for TransGTR, MetaST, MMDNet, and FusionTN. This efficiency advantage arises because only the lightweight prompt network and prediction head are fine-tuned during adaptation, while the memory pool and spatial mapping modules remain frozen. TransMod also achieves competitive inference latency. Instead of re-computing the full station-level graph attention at test time, it performs a memory lookup followed by zone-level propagation, resulting in sub-10\,ms inference latency. In terms of scalability, memory retrieval has complexity $\mathcal{O}(N_z \cdot M \cdot d)$, where the number of prototypes is fixed at $M{=}128$ regardless of city scale. Therefore, the inference cost grows linearly with the number of zones and remains tractable for dense metropolitan grids with thousands of zones.

\subsection{Ablation Studies}

To systematically examine the internal effectiveness and stability of TransMod, we conduct ablation studies by selectively removing key components from the full framework. As shown in Figure~\ref{fig:my_label3}(\subref{fig:subfig_a3}) and (\subref{fig:subfig_b3}), we construct four ablated variants: \textbf{w/o GE}, which removes Gaussian uncertainty embeddings and uses deterministic station representations; \textbf{w/o SA}, which replaces the learnable soft assignment mechanism with hard nearest-zone assignment; \textbf{w/o CT}, which removes the cross-modal alignment losses, including MMD and InfoNCE, and trains source modalities without representation alignment; and \textbf{w/o MT}, which replaces the memory-based transfer module with a standard GRU decoder. All ablated variants show consistent performance degradation across NYC and Chicago on all evaluation metrics, confirming that each component contributes to the overall effectiveness of TransMod. A closer examination further reveals the distinct role of each component. 

Among all variants, removing the soft assignment mechanism leads to the most pronounced degradation, with MAE increasing by 13.0\% in NYC and 16.0\% in Chicago. This result indicates that hard nearest-zone assignment can distort the aggregation of station-level signals into zone-level representations, whereas the learnable and geography-regularized soft assignment is essential for bridging the spatial granularity mismatch between station-based and zone-based systems. Removing Gaussian uncertainty embeddings (\textbf{w/o GE}) and cross-modal alignment (\textbf{w/o CT}) also causes moderate performance drops. Without uncertainty modeling, TransMod becomes less capable of suppressing unreliable auxiliary signals; without cross-modal alignment, bike-sharing and metro representations remain distributionally inconsistent, which weakens the quality of temporal patterns stored in the memory pool. Finally, replacing the memory-based transfer module with a GRU decoder (\textbf{w/o MT}) increases MAE by 8.8\% in NYC and 11.8\% in Chicago. Although the GRU decoder can still capture periodic temporal patterns from source data, its weaker performance suggests that the memory pool encodes richer zone-structure-conditioned temporal dynamics beyond simple recurrent statistics. This structured prior is particularly important when target-domain temporal observations are limited. 

\subsection{Robustness Analysis}

\begin{figure}[t]
    \centering
    \includegraphics[width=\linewidth]{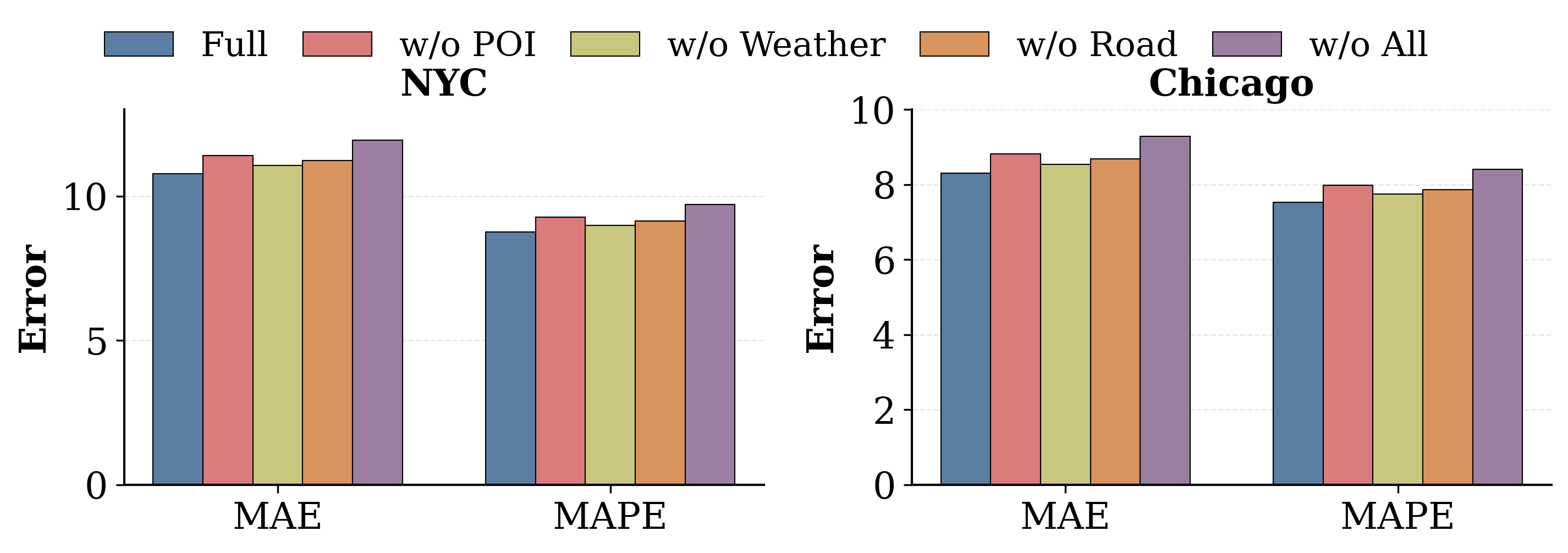}
    \caption{Performance under auxiliary feature removal on NYC and Chicago.}
    \label{fig:my_lable5}
    \Description{}
\end{figure}

We conduct a comprehensive robustness analysis of TransMod from four perspectives: cross-city transfer, cross-modal transfer, practical perturbations, and auxiliary feature degradation. The goal is to examine whether TransMod maintains reliable forecasting performance under diverse data-limited and heterogeneous settings.

\noindent\textbf{Cross-city and cross-modal transfer.}~To assess cross-city transferability, we examine whether TransMod can reuse mobility knowledge learned from one city when adapted to another under limited target supervision. Specifically, TransMod is pre-trained on New York City and fine-tuned on Chicago with varying proportions of ride-hailing (RH) data. We compare three settings: \textit{No Transfer}, which trains from scratch using only Chicago RH data; \textit{With Transfer}, which initializes TransMod with the NYC-pre-trained memory pool; and \textit{Upper Bound}, which uses the full Chicago RH dataset. As shown in Figure~\ref{fig:my_label3}(\subref{fig:subfig_a4}) and (\subref{fig:subfig_b4}), TransMod consistently outperforms No Transfer and approaches the Upper Bound when target-city data are limited. With only 10\% of Chicago data, TransMod achieves an RMSE of 12.38, compared with 16.21 without transfer, indicating that the source-learned memory provides effective temporal priors for target-city adaptation.

We further evaluate cross-modal transferability using two target tasks: ride-hailing demand prediction and bike-sharing demand prediction. For the RH target, auxiliary sources include bike-sharing, metro, or both; for the bike-sharing target, sources include metro, ride-hailing, or both. This setting tests whether TransMod can generalize beyond a single predefined source--target direction. As shown in Figure~\ref{fig:my_label3}(\subref{fig:subfig_a5}) and (\subref{fig:subfig_b5}), TransMod achieves the lowest MAPE across all six source--target configurations and maintains more stable performance than competing baselines. These results confirm that TransMod supports robust knowledge transfer across both cities and mobility modes under data-limited conditions.

\noindent\textbf{Robustness to practical perturbations.}~We also evaluate TransMod under spatial coarsening, data scarcity, and long-horizon forecasting. Under spatial coarsening, TransMod remains stable as the grid size increases from 0.5 km to 2 km, with RMSE rising by only 9.5\%, compared with 17\%--24\% for competing baselines. Under data scarcity, TransMod consistently outperforms all baselines and achieves performance comparable to methods trained with 70\%--100\% of the data using only 30\% of RH records, demonstrating its ability to exploit auxiliary mobility signals when target observations are sparse. For long-horizon forecasting, TransMod remains the most stable from $t$ to $t+12$, with RMSE increasing by 28.3\%, compared with 40\%--50\% for most baselines. These results show that TransMod effectively exploits transferable spatio-temporal structure across heterogeneous mobility modes, enabling reliable forecasting under spatial mismatch, limited target data, and extended prediction horizons.

\noindent\textbf{Robustness to auxiliary feature degradation.}~Finally, we evaluate the impact of removing each auxiliary feature category by replacing its inputs with zero vectors during both pre-training and adaptation. As shown in Figure~\ref{fig:my_lable5}, removing POI distributions causes the largest degradation, increasing MAE by 5.8\% in NYC and 6.1\% in Chicago. This indicates that POI features play a critical role in both the soft assignment prior and the prompt network. Removing road-network features results in moderate performance drops of 4.3\% and 4.7\%, while meteorological features have the smallest individual impact, with MAE increases of 2.6\% and 2.9\%. When all auxiliary features are removed simultaneously, MAE increases by 10.8\% in NYC and 11.8\% in Chicago. This degradation remains non-catastrophic, suggesting that the Gaussian uncertainty mechanism improves robustness against missing or corrupted auxiliary inputs.

\subsection{Hyperparameter Sensitivity Analysis}

\begin{figure}[t]
  \centering
  \includegraphics[width=\linewidth]{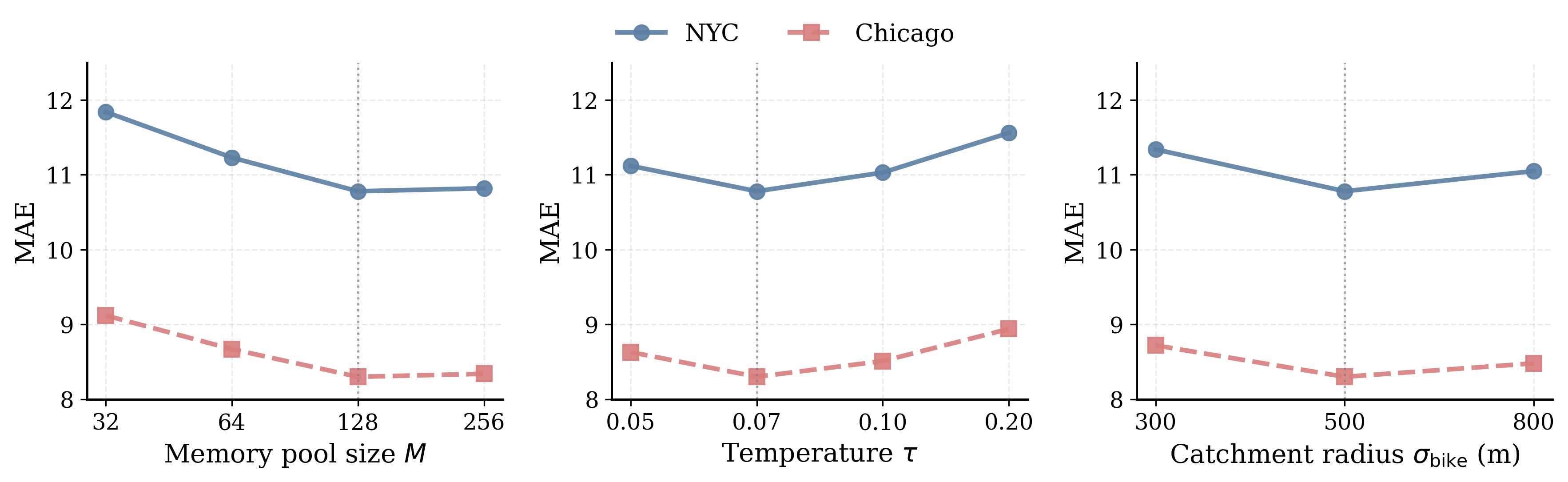}
  \caption{Hyperparameter sensitivity of TransMod on NYC and Chicago.}
  \label{fig:my_label4}
  \Description{Hyperparameter sensitivity results for memory pool size, InfoNCE temperature, and bike-sharing bandwidth on NYC and Chicago.}
\end{figure}

We analyze the sensitivity of TransMod to three key hyperparameters: memory pool size~$M$, InfoNCE temperature~$\tau$, and the mobility-specific bandwidth $\sigma_{\text{bike}}$ for bike-sharing, which controls the geographic decay in the soft assignment prior defined in Eq.~\ref{eq:soft_assign_geo}. In each experiment, one hyperparameter is varied within a predefined range while the others are fixed at their default values, i.e., $M{=}128$, $\tau{=}0.07$, and $\sigma_{\text{bike}}{=}500\,m$. The default setting is marked by the dotted vertical line in Figure~\ref{fig:my_label4}. We report MAE on both NYC and Chicago to assess cross-city consistency.

\noindent\textbf{Memory pool size $M$.}~Increasing $M$ from 32 to 128 reduces MAE from 11.84 to 10.78 on NYC and from 9.12 to 8.30 on Chicago, confirming that a larger memory pool provides more diverse temporal prototypes. However, further increasing $M$ brings negligible improvement and slightly increases MAE to 10.82 and 8.34 on NYC and Chicago, respectively. We therefore set $M{=}128$ as the default to balance prototype coverage and memory cost.

\noindent\textbf{InfoNCE temperature $\tau$.}~TransMod is most sensitive to $\tau$ among the three hyperparameters. A small temperature, e.g., $\tau{=}0.05$, makes memory retrieval overly concentrated on dominant prototypes, suppressing useful secondary temporal patterns and increasing NYC MAE to 11.12. In contrast, a large temperature, e.g., $\tau{=}0.20$, produces overly uniform attention weights, weakening the discriminative structure of the retrieved memory and increasing NYC MAE to 11.56. The default value $\tau{=}0.07$ achieves the best balance between prototype specificity and coverage across both cities.

\noindent\textbf{Bandwidth $\sigma_{\text{bike}}$.}~Compared with $M$ and $\tau$, TransMod is less sensitive to $\sigma_{\text{bike}}$. Reducing $\sigma_{\text{bike}}$ to 300\,m makes the assignment overly localized and closer to hard nearest-zone mapping, raising NYC MAE from 10.78 to 11.34. Increasing it to 800\,m introduces a milder degradation, with NYC MAE increasing to 11.05, suggesting that overly broad kernels may introduce noise from distant zones. The stable performance between 500\,m and 800\,m indicates that the soft assignment mechanism is robust to moderate variations in the geographic prior. For metro, we use a larger default bandwidth $\sigma_{\text{metro}}{=}800\,m$, reflecting the wider catchment area of rail stations.

%% file: 04_Related_Work.tex
\section{Related Work}

\subsection{Urban Mobility Demand Forecasting}

Demand prediction is a core problem in urban mobility systems, such as bike-sharing, ride-hailing, and public transit. Existing studies have extensively explored deep learning models to capture temporal dynamics and spatial correlations in mobility demand, incorporating external factors such as weather, POIs, and land-use information~\cite{10.5555/3298239.3298479,10.5555/3504035.3504351,wang2020deep,zhou2024bike}. These methods have significantly improved forecasting accuracy by modeling complex spatio-temporal dependencies in urban travel behavior. However, most approaches achieve this within a single mobility system, where the spatial units, temporal observations, and demand generation processes are assumed to be internally consistent~\cite{liang2023deep,liang2024time,behroozi2025predicting}. Such an assumption becomes restrictive in multi-modal urban environments, where different systems may observe the same underlying mobility demand through different spatial abstractions, such as stations, stops, or zones. Consequently, models trained for one system often cannot be directly transferred to another without resolving the mismatch in representation and data availability. This design limits their ability to exploit transferable demand patterns between data-rich and data-scarce mobility systems. Even recent multi-modal studies often rely on independent models or simple feature fusion, treating auxiliary modes mainly as additional inputs rather than transferable knowledge sources~\cite{jin2022selective}. As a result, they remain insufficient for forecasting across heterogeneous mobility systems with mismatched spatial representations and uneven temporal availability.

\subsection{Spatio-Temporal Representation Learning}
Learning effective representations has become central to transportation forecasting, as it enables models to capture spatial dependencies, temporal dynamics, and their interactions in traffic and mobility networks. Early approaches integrate graph-based spatial modeling with recurrent or convolutional temporal modules to learn structured spatio-temporal representations, such as DCRNN, STGCN, and Graph WaveNet~\cite{li2017diffusion,yu2017spatio,10.5555/3367243.3367303}. These models represent transportation systems as graphs, where nodes denote sensors, stations, or regions, and edges encode spatial proximity or mobility interactions. This graph-based formulation provides a powerful way to model localized dependencies, but it also ties the learned representation to a specific node definition and spatial partition. Subsequent studies enhance modeling flexibility by learning adaptive graph structures and node embeddings directly from data, capturing non-stationary spatial correlations beyond predefined adjacency matrices~\cite{bai2020adaptive,zhang2019spatial}. Such adaptive representations improve intra-system forecasting, especially when spatial correlations evolve over time, but they still assume that the source and target observations share compatible spatial semantics. More recent works further explore attention-based and decoupled spatial--temporal architectures to capture long-range dependencies~\cite{Zheng_Fan_Wang_Qi_2020}. Despite these advances, most methods are designed for single systems with fixed spatial units, where the node set and observation structure remain consistent during training and inference. This assumption limits their applicability to heterogeneous mobility systems with mismatched spatial granularity and uneven temporal availability, motivating a unified representation space for cross-modal knowledge transfer.

\subsection{Knowledge Transfer Across Modes}
Transfer learning and domain adaptation have been widely studied in urban mobility forecasting to alleviate data scarcity by transferring knowledge across cities, regions, or transportation systems~\cite{10.5555/3367243.3367301,10.1145/3447548.3467330,ijcai2022p282,huang2023traffic,wang2024ride}. Existing approaches typically reuse spatio-temporal patterns learned from data-rich source domains to improve forecasting in data-scarce target domains, and are often grounded in domain adaptation principles such as distribution matching, adversarial learning, and feature-level invariance. These methods are effective when the source and target domains share comparable spatial units and observation structures, since the transferred representations can be aligned within a relatively consistent feature space. Recent studies further extend these ideas to graph-based mobility forecasting by aligning node embeddings, edge structures, or spatio-temporal graph representations between source and target domains~\cite{ganin2016domain,tzeng2017adversarial,jin2023transferable,wang2024stone,hu2024prompt,yang2025cross}. However, graph-based alignment still commonly relies on structural compatibility across domains, including similar node definitions, graph topologies, and observation patterns. This reliance becomes restrictive in cross-modal mobility transfer, where station-based and zone-based systems differ not only in spatial representation, but also in temporal availability and data generation processes. As a result, existing transfer methods remain limited when transferring knowledge across heterogeneous mobility modes, such as bike-sharing, metro, and ride-hailing systems~\cite{wang2024fusiontransnet,jiang2025leveraging,hua2025transfer}.

%% file: 05_Conclusion.tex
\section{Conclusion and Future Work}

In this paper, we propose TransMod, a unified framework for urban mobility forecasting across heterogeneous modes with mismatched spatial granularity and uneven temporal observability. TransMod builds a shared spatial representation to bridge station-based and zone-based systems, aligns cross-modal representations, and retrieves transferable temporal patterns from data-rich source modes through memory-based transfer. Experiments on New York City and Chicago datasets show that TransMod outperforms state-of-the-art baselines and remains robust under spatial coarsening, data scarcity, cross-city transfer, and long-horizon forecasting. These results highlight the effectiveness of cross-modal spatio-temporal knowledge transfer in data-heterogeneous urban settings, offering a scalable step toward generalizable urban mobility forecasting.

Future work may extend TransMod in several directions. Incorporating additional mobility systems, such as shared scooters, on-demand buses, and autonomous fleets, would further test its generality under broader urban heterogeneity. Online or continual memory updates could improve adaptability to evolving mobility patterns and sudden demand shifts. Multi-city pre-training may learn more general temporal priors and reduce adaptation costs for newly instrumented cities. Making memory prototypes easier to understand could further improve the practical value of memory-augmented transfer for urban planning applications. 